\pdfoutput=1

\documentclass[11pt]{article}

\usepackage{acl}

\usepackage{times}
\usepackage{latexsym}

\usepackage[T1]{fontenc}

\usepackage[utf8]{inputenc}

\usepackage{microtype}

\usepackage{inconsolata}
\usepackage{todonotes}
\usepackage{rotating}
\usepackage{graphicx}
\usepackage{balance} 
\usepackage{lscape}
\usepackage{amsmath}
\usepackage{amssymb}
\usepackage{amsthm}
\usepackage{float}
\usepackage{multirow}
\usepackage{algorithm}
\usepackage{algpseudocode}
\usepackage{tcolorbox}
\usepackage{booktabs}
\usepackage{pifont}
\usepackage{bbm}
\usepackage[normalem]{ulem}
\usepackage{hyperref}
\usepackage{algorithmicx}
\useunder{\uline}{\ul}{}

\newcommand{\EntitySet}{E}
\newcommand{\PredicateSet}{R}
\newcommand{\triple}{\langle h,r,t\rangle}
\newcommand{\quadruple}{\langle h,r,t,c\rangle}
\newcommand{\model}{M_{\theta}}
\newcommand{\PosTripleSet}{\mathcal{T}}
\newcommand{\NegTripleSet}{\mathcal{T}^-}

\newcommand{\PredictionSet}[1]{\Gamma_{\scriptscriptstyle \mathrm{#1}}^\alpha}
\newcommand{\CPSet}{\PredictionSet{CP}}
\newcommand{\ICPSet}{\PredictionSet{ICP}}
\newcommand{\UnKGCPSet}{\PredictionSet{UnKGCP}}
\newcommand{\ComICPSet}{\Gamma_{\scriptscriptstyle \mathrm{ICP}}^{\alpha'}}
\newcommand{\FisherSet}{\PredictionSet{FPI}}
\newcommand{\QuantileSet}{\PredictionSet{QR}}

\newcommand{\NamedTSet}[1]{\mathcal{T}_\mathrm{#1}}
\newcommand{\TrainingSet}{\NamedTSet{train}}

\newcommand{\CalibrationSet}{\NamedTSet{cal}}
\newcommand{\TestSet}{\NamedTSet{test}}

\newcommand{\TrainingTime}{T_\mathrm{train}}
\newcommand{\InferenceTime}{T_\mathrm{infer}}

\newcommand{\SemiTripleSet}{\mathcal{T}^{\text{semi}}}
\newcommand{\PSLTripleSet}{\mathcal{T}^{\text{psl}}}

\algdef{SE}[DOWHILE]{Do}{doWhile}{\algorithmicdo}[1]{\algorithmicwhile\ #1}%

\newtheorem{theorem}{Theorem}

\newtheorem{proposition}{Proposition}
\newtheorem{remark}{Remark}

\algnewcommand{\LineComment}[1]{\State \(\triangleright\) #1}
\title{ Certainty in Uncertainty: Reasoning over Uncertain Knowledge Graphs with Statistical Guarantees }

\author{%
Yuqicheng Zhu\textsuperscript{1,2}\thanks{Equal contribution.}, 
Jingcheng Wu\textsuperscript{1}\footnotemark[1],
Yizhen Wang\textsuperscript{1},  
Hongkuan Zhou\textsuperscript{1,2},\\
\textbf{Jiaoyan Chen\textsuperscript{3},}
\textbf{Evgeny Kharlamov\textsuperscript{2,4},} \textbf{Steffen Staab\textsuperscript{1,5}}
\\
\textsuperscript{1}University of Stuttgart, 
\textsuperscript{2}Bosch Center for AI, 
\\
\textsuperscript{3}The University of Manchester, 
\textsuperscript{4}University of Oslo, 
\textsuperscript{5}University of Southampton\\
\texttt{yuqicheng.zhu@de.bosch.com}\\
}

\begin{document}
\maketitle
\begin{abstract}
Uncertain knowledge graph embedding (UnKGE) methods learn vector representations that capture both structural and uncertainty information to predict scores of unseen triples.
However, existing methods produce only point estimates, without quantifying predictive uncertainty—limiting their reliability in high-stakes applications where understanding confidence in predictions is crucial.
To address this limitation, we propose \textsc{UnKGCP}, a framework that generates prediction intervals guaranteed to contain the true score with a user-specified level of confidence. 
The length of the intervals reflects the model’s predictive uncertainty.
\textsc{UnKGCP} builds on the conformal prediction framework but introduces a novel nonconformity measure tailored to UnKGE methods and an efficient procedure for interval construction.
We provide theoretical guarantees for the intervals and empirically verify these guarantees.
Extensive experiments on standard benchmarks across diverse UnKGE methods further demonstrate that the intervals are sharp and effectively capture predictive uncertainty. To support future research on this topic, we release our code\footnote{\url{https://github.com/0sidewalkenforcer0/UnKGCP}}.

\end{abstract}

\section{Introduction}
Knowledge graphs (KGs) represent factual knowledge as triples of the form $\langle$ \textit{Head Entity}, \textit{Predicate}, \textit{Tail Entity}$\rangle$, capturing relationships between real-world entities \cite{hogan2021knowledge}.
Knowledge in KGs can be uncertain due to noise and errors from inaccurate automated extraction processes \cite{pujara2013knowledge}, or because some facts are inherently probabilistic, such as molecular interactions \cite{szklarczyk2016string}. To capture such uncertainty, uncertain KGs (UnKGs) associate each triple with a score that reflects the likelihood of the fact being true \cite{wu2012probase, speer2017conceptnet, mitchell2018nell}.

Reasoning over UnKGs aims to predict the score of unseen triples, leveraging the structure and uncertainty information encoded in the observed graph. 
Existing approaches \cite{chen2019embedding, chen2021passleaf, chen2021probabilistic, kgv} extend KG embedding (KGE) techniques to UnKGs, which we refer to as UnKGE methods. Specifically, these methods represent entities and predicates as numerical vectors, assess the plausibility of triples based on distance \cite{bordes2013translating} or dot product \cite{nickel2011three}, and then map this plausibility to a score in the range $[0,1]$. 

However, existing UnKGE models produce only point estimates without capturing how confident the model is in its predictions.
In real-world applications~\cite{FM4SU, MultiADS}, especially in high-stakes domains, it is crucial to know the range within which the true score is likely to fall.
For example, when predicting the likelihood of a harmful drug interaction based on a biomedical KG, a point estimate of $0.3$ might suggest low risk. However, if the model also indicated that plausible scores range from 0.2 to 0.95, it would reveal high uncertainty, indicating that further investigation is needed before taking clinical action.

To the best of our knowledge, no existing method provides a statistically grounded way to quantify uncertainty in the predictions of UnKGE methods. We take the first step toward addressing this gap by introducing \textsc{UnKGCP}, a framework that applies \emph{conformal prediction} \cite{vovk2005algorithmic} to quantify uncertainty through a prediction interval—a set of plausible values that is guaranteed to contain the ground truth with a user-specified confidence level.
The core idea is to assess how "atypical" a candidate prediction is compared to previously seen data using a \emph{nonconformity score}. Based on these scores, the method selects a threshold that ensures the constructed interval includes the ground truth with the desired level of confidence.
Specifically, we introduce \textbf{a novel nonconformity measure tailored to UnKGE methods} that allows the prediction intervals to adapt to the difficulty of each query, along with \textbf{an efficient procedure for constructing such intervals}. 

We provide theoretical guarantees on the coverage of the ground truth by the prediction intervals (Proposition \ref{prop:main}) and validate our approach through extensive experiments on commonly used UnKG benchmarks across a range of UnKGE methods.
Our empirical study shows that: 
(1) \textsc{UnKGCP} produces prediction intervals that both satisfy the theoretical guarantees and remain sharp and informative;
(2) the intervals adapt to query-specific uncertainty;
(3) \textsc{UnKGCP} is sample-efficient, achieving similar performance using only about 20\% of the calibration set.

\section{Related Work}
\textbf{UnKGE Methods.} Several UnKGE methods have been proposed to support reasoning under triple-level uncertainty \citep{chen2019embedding, chen2021passleaf, chen2021probabilistic}. As the first in this line of research, UKGE \citep{chen2019embedding} extends DistMult \citep{yang2015distmult} to UnKGs by mapping plausibility scores to the $[0,1]$ range, replacing the loss function with mean squared error, and augmenting training data using probabilistic soft logic. PASSLEAF \citep{chen2021passleaf} generalizes this framework to support a broader range of KGE backbones and improves negative sampling by predicting scores for negative triples using semi-supervised learning. In contrast to vector-based models, \citet{chen2021probabilistic} represent entities as boxes and encode relations as affine transformations between boxes, achieving improved performance and robustness to noise.
Another line of work addresses reasoning under schema-level uncertainty. For example, \citet{zhu2023towards, zhu2024approx} approximate probabilistic inference in statistical $\mathcal{EL}$ using box embeddings.

\noindent
\textbf{Uncertainty Quantification in KGE.} As highlighted by \citet{zhu2024predictive}, the predictions of KGE models can vary substantially with minor changes to training conditions (e.g., random seed), underscoring the importance of uncertainty quantification in KGE. Some recent work has explored this: \citet{Tabacof2020calibration, Safavi2020calibration} apply post-hoc calibration techniques to map plausibility scores from KGEs into calibrated probabilities. 
However, uncertainty quantification in the context of UnKGE has been largely overlooked. \citet{zhu2024approx} produce intervals via ensemble methods, but these intervals lack formal statistical guarantees.

\noindent
\textbf{Conformal Prediction.} This work applies conformal prediction, a general framework for uncertainty quantification with finite-sample statistical guarantees. Conformal prediction has been applied across various domains, including image classification \citep{Angelopoulos2021image}, natural language processing \citep{maltoudoglou2020bert, campos2024conformal}, node classification and regression on graphs \citep{huang2024uncertainty, zargarbashi2023conformal, zargarbashi2023inductive}, and link prediction on deterministic KGs \citep{zhu2025conformalized, zhu2025condkgcp}.

Among these, \citet{zhu2025conformalized} and \citet{zhu2025condkgcp} are most closely related to our work, but they focus on link prediction in deterministic KGs and adopts nonconformity measures and set construction procedure specifically designed for that context. 
These methods cannot be directly applied to UnKGE tasks due to fundamental differences in output space and objectives: link prediction involves ranking a finite set of candidate entities and yields discrete prediction sets to quantify uncertainty, whereas score prediction in UnKGE requires constructing real-valued prediction intervals for continuous outputs. As a result, the design of nonconformity scores, set construction, and theoretical guarantees differs substantially.

\section{Preliminaries}

\subsection{Uncertain Knowledge Graph Embeddings}
Let $\EntitySet$ and $\PredicateSet$ represent finite sets of \emph{entities} and \emph{predicates}, respectively. A KG is a subset of $\EntitySet\times\PredicateSet\times\EntitySet$, where each element, called \emph{triple}, represents a fact.
An UnKG extends KG by associating each fact with a confidence score indicating the likelihood of the fact being true.
Formally, an UnKG can be defined as a set of \emph{weighted triples}:
\begin{equation*}
    \big\{\quadruple \mid \triple\in\EntitySet\times\PredicateSet\times\EntitySet, c\in[0,1]\big\}.
\end{equation*}

An UnKGE model is a function $\model:\EntitySet\times\PredicateSet\times\EntitySet\rightarrow[0,1]$ that assigns confidence scores to triples. 
The parameters $\theta$ of the model are learned by minimizing the discrepancy between predicted and ground truth confidence scores. 
A typical training objective is the mean squared error \cite{chen2019embedding}:
\begin{equation}
    L = \sum_{(q,c)\in\PosTripleSet\cup\NegTripleSet}|\model(q)-c|^2,
\end{equation}
where $q = \langle h, r, t \rangle$ is a query triple, and, $\PosTripleSet, \NegTripleSet$ denote the sets of positive and negative training examples, respectively.

Note that UnKGs often do not contain explicit negative examples. Negative triples are commonly generated by corrupting positive triples, for example, by replacing the head or tail entity with a randomly selected entity from $\EntitySet$ \cite{chen2019embedding}. 
However, confidence scores for these negative triples are not simply assigned a value of 0. Instead, \citet{chen2019embedding} employ probabilistic soft logic to estimate their scores, while \citet{chen2021passleaf} adopt a semi-supervised learning framework for this purpose.


\subsection{Conformal Prediction}
In this section, we recall essential concepts from conformal prediction as introduced in \citet{vovk2005algorithmic}.
Consider a dataset $Z=\{(x_i, y_i)\}_{i=1}^n$, 
with inputs $x_i\in\mathcal{X}$ and the corresponding labels $y_i \in \mathcal{Y}$.
We denote the space of individual examples by $\mathcal{Z}=\mathcal{X}\times\mathcal{Y}$, and the space of all possible example sets by $\mathcal{Z}^*$.

\subsubsection{Confidence Predictor} \label{sec:conf_pred}
Given a test input $x_{n+1}$, our goal is to design an algorithm $\Gamma$ that, instead of predicting a single label for $y_{n+1}$, outputs a \emph{prediction interval}—a subset of $\mathcal{Y}$ that contains the true label with a specified \emph{confidence level} $\alpha\in[0,1]$.
To reflect the trade-off between confidence and informativeness, 
the prediction intervals are required to expand as $\alpha$ increases: intuitively, achieving higher confidence necessitates including more possible labels.

Formally, a \emph{confidence predictor} is a measurable function
\begin{equation}
    \Gamma:\mathcal{Z}^*\times\mathcal{X}\times[0,1]\rightarrow2^{\mathcal{Y}},
\end{equation}
that maps a set of (training) examples, a test input, and a desired confidence level to a subset of possible labels. 
In our case, this subset corresponds to an interval.
For notational convenience, we write $\Gamma^\alpha(Z,x):=\Gamma(Z,x,\alpha)$ to denote the prediction interval at level $\alpha$.
We require the confidence predictor to satisfy the following \emph{monotonicity property}:
\begin{equation}\label{eq:inter_prop}
    \Gamma^{\alpha_1}(Z,x_{n+1})\subseteq\Gamma^{\alpha_2}(Z,x_{n+1}), \forall\alpha_1\leq\alpha_2.
\end{equation}

The quality of a confidence predictor is evaluated based on three key \textbf{desiderata}: \emph{validity}, \emph{efficiency}, and \emph{conditionality}.
\begin{itemize}
    \item \textbf{Validity} ensures that, in the long run, the prediction interval covers the true label with probability at least $\alpha$.
    Formally, 
    the coverage of a confidence predictor $\Gamma$ at level $\alpha$ is defined as 
    \begin{align}\label{eq:coverage}
        &Cov(\Gamma^\alpha):=\\
        &\notag\underset{\small\substack{Z \sim \mathcal{P}^n, \\ (x_{n+1}, y_{n+1}) \sim \mathcal{P}}}{\mathbb{P}}
        \Big(y_{n+1}\in\Gamma^\alpha(Z,x_{n+1})\Big),
    \end{align}
    where $\mathcal{P}$ denotes the (unknown) joint distribution over examples.
    We say that $\Gamma$ is \emph{exactly valid} if $Cov(\Gamma^\alpha)=\alpha$, and \emph{conservatively valid} if $Cov(\Gamma^\alpha)\geq\alpha$.
    \item \textbf{Efficiency} refers to the tightness of the prediction intervals. Given the same confidence level, a more efficient confidence predictor produces sharper (i.e., more informative) prediction intervals.
    \item \textbf{Conditionality} expresses the degree to which the confidence predictor adapts to the difficulty of individual examples. Ideally, the size of the prediction interval should reflect how uncertain the model is about the specific input $x_{n+1}$: smaller for easy cases and larger for hard ones.    
\end{itemize}
 
\subsubsection{Conformal Predictor}\label{sec:cp}
A \emph{conformal predictor} is a confidence predictor that provides rigorous validity guarantees.
It leverages a \emph{nonconformity measure} $S:\mathcal{Z}^*\times\mathcal{Z\rightarrow\mathbb{R}}$, which quantifies how "strange" a test example appears relative to observed examples. 
Given a model $\hat f_{Z}:\mathcal{X}\rightarrow\mathcal{Y}$ trained on $Z$, a common nonconformity measure for regression tasks is the absolute residual:
\begin{equation}\label{eq:absolute_residual}
    S\big(Z, (x,y)\big)=\left|\hat f_{Z}(x)-y\right|.
\end{equation}
Applying $S$ to an example $z$ yields a \emph{nonconformity score} $s=S(Z,z)$, which reflects how atypical $z$ appears when compared against the examples in $Z$.

To generate a prediction interval for a test input $x_{n+1}$, the conformal predictor proceeds as follows.
For each candidate $y \in \mathcal{Y}$, it forms an augmented dataset $Z' = Z \cup {(x_{n+1}, y)}$ and computes nonconformity scores for all examples in $Z'$. 
In particular, it computes
\begin{align}
    & s_i:=S(Z', z_i), i=1,\dots, n,\\
    \notag& s_{n+1}:=S\big(Z', (x_{n+1}, y)\big).
\end{align}
The label $y$ is included in the prediction interval if its nonconformity score $s_{n+1}$ is not among the largest $1-\alpha$ fraction of scores in $Z'$, that is:
\begin{align}\label{eq:full_set_construct}
    &\CPSet(Z, x_{n+1}):=\Big\{y\in\mathcal{Y}:\\
    \notag&\frac{|\{i=1,\dots,n+1:s_i\geq s_{n+1}\}|}{n+1}>1-\alpha\Big\}.
\end{align}

By constructing prediction intervals as described above, all conformal predictors have the following validity guarantees.
\begin{theorem}[\citet{vovk2005algorithmic}, \citet{lei2018distribution}]\label{th:validity}
    Assume the examples in $Z$ and the test example $z_{n+1}$ are independent and identically distributed (i.i.d). 
    For any confidence level $\alpha\in[0,1]$ and any nonconformity measure $S$, the conformal predictor $\CPSet$ is conservatively valid:
    \begin{equation}
        \mathbb{P}\big(y_{n+1}\in\CPSet(Z,x_{n+1})\big)\geq\alpha.
    \end{equation}
    Furthermore, if $\{s_i\}_{i=1}^n$ contains no ties, $\CPSet$ is also asymptotically exactly valid:
    \begin{equation}
        \lim_{n\rightarrow\infty}\mathbb{P}\big(y_{n+1}\in\CPSet(Z,x_{n+1})\big)=\alpha.
    \end{equation}
\end{theorem}
\begin{remark}
    The validity guarantees of conformal prediction hold under the even weaker assumption of \emph{exchangeability} \cite{shafer2008tutorial, vovk2005algorithmic}. Exchangeability allows dependencies among examples, as long as their joint distribution remains invariant under permutations.
\end{remark}

\section{Conformalized Uncertain Knowledge Graph Embeddings (UnKGCP)}
Conformal prediction is a general uncertainty quantification framework requiring careful adaptation for specific tasks via tailored nonconformity measures and efficient prediction interval constructions.
In this section, 
we introduce \textbf{an efficient way to construct prediction intervals}, analyse its time complexity and prove its validity guarantees. Moreover, we propose \textbf{a novel nonconformity measure} designed for UnKGE models, ensuring query-specific prediction intervals.

\subsection{Problem Setup}
We consider a set of weighted triples $\mathcal{T}=\{tr_i\}_{i=1}^n$, 
where each $tr_i=(q_i, c_i)$ consists of a query triple $\langle h,r,t\rangle$ and an associated confidence score $c_i\in[0,1]$.
Given an UnKGE model $M_{\mathcal{T}}$ trained on $\mathcal{T}$, the reasoning task on UnKGs is to predict the confidence score for a test query $q_{n+1}$.
We aim to quantify the uncertainty in model predictions by constructing prediction intervals $\Gamma^{\alpha}(\mathcal{T}, q_{n+1})$ at a use-specified confidence level $\alpha\in[0,1]$, satisfying the properties 
as described in Section~\ref{sec:conf_pred}.

\subsection{Efficient Set Construction}
Applying conformal prediction as described in Section \ref{sec:cp} to UnKGE requires examining infinitely many potential confidence scores $c\in[0,1]$ and training a new UnKGE model for each query-potential value pair $(q_{n+1}, c)$, which is computationally prohibitive.

To overcome this, we employ inductive conformal prediction (ICP) \cite[Section 4.2][]{vovk2005algorithmic, lei2018distribution} to construct prediction intervals efficiently and avoid examining infinitely many cases.
Specifically, we randomly partition $\mathcal{T}$ into two disjoint sets: a \emph{proper training set} $\TrainingSet=\{tr_i\}_{i=1}^{m}$ and a \emph{calibration set} $\CalibrationSet=\{tr_i\}_{i=m+1}^{n}$ of size $\ell=n-m$. An UnKGE model is trained exclusively on $\TrainingSet$, after which it remains fixed to compute nonconformity scores on $\CalibrationSet$ and new queries.

Given a nonconformity measure $S$ and a user-specified confidence level $\alpha\in[0,1]$, the ICP-based prediction interval for a test query $q_{n+1}$ is defined as 
\begin{align}\label{eq:split_set_construct}
    &\ICPSet(\mathcal{T}, q_{n+1}):=\Big\{c\in[0,1]:\\
    \notag&\frac{|\{i=m+1,\dots,n+1:s_i\geq s_{n+1}\}|}{\ell+1}>1-\alpha\Big\},
\end{align}
where 
\begin{align}
    & s_i:=S\big(\TrainingSet, tr_i\big), i=m+1,\dots, n,\\
    \notag& s_{n+1}:=S\big(\TrainingSet, (q_{n+1}, c)\big),
\end{align}

\subsubsection{Time Complexity}
This procedure avoids repeated model retraining and significantly improves computational efficiency.
Given $k$ test queries, the overall computational complexity scales as
\begin{equation}
    \mathcal{O}\Big(\TrainingTime+(\ell+k)\InferenceTime+\ell\log \ell+k\log \ell\Big),
\end{equation}
where $\TrainingTime$ is the one-time cost of training the UnKGE model. For mainstream UnKGE methods $\TrainingTime=\mathcal{O}(|E|d)$, with $d$ denoting the embedding dimension. 
$\InferenceTime=\mathcal{O}(d)$ is the time to compute the confidence score for a query.
We allocate time $\ell\log\ell$ to sort the nonconformity scores obtained from the calibration set $\CalibrationSet$, $\log\ell$ to determine the rank of $s_{n+1}$ among the scores in $\{s_i\}_{i=m+1}^n$. 
Since only $k\InferenceTime = \mathcal{O}(kd)$ and $k\log\ell$ depend on $k$, while $\TrainingTime$, $\ell\InferenceTime$, and $\ell\log\ell$ are constant in $k$, the overall complexity as $k \to \infty$ is $\mathcal{O}\big(k(d + \log \ell)\big)$—\textbf{asymptotically linear} in $k$.

When scaling to larger graphs, the dominant cost is $\TrainingTime$. 
Since $\InferenceTime$ scales only linearly with the embedding dimension $d$. All other terms are completely independent of the graph size. Thus, once the UnKGE model is trained, \textbf{our method runs with a computational cost that is agnostic to the graph size}.

\subsubsection{Validity Guarantees}
We show that the ICP-based conformal predictor retains the formal validity guarantees stated in Theorem~\ref{th:validity}, while benefiting from a more efficient set construction process. The proof is provided in Appendix~\ref{app:proof}.
\begin{proposition} \label{prop:main}
    Assume weighted triples in $\mathcal{T}$ and the test weighted triple $(q_{n+1}, c_{n+1})$ are i.i.d. For any confidence level $\alpha\in[0,1]$ and any nonconformity measure $S$, the ICP-based conformal predictor $\ICPSet$ is conservatively valid:
    \begin{equation}
        \mathbb{P}\big(c_{n+1}\in\ICPSet(\mathcal{T}, q_{n+1})\big)\geq\alpha.
    \end{equation}
    Furthermore, if $\{s_i\}_{i=m+1}^n$ contains no ties, $\ICPSet$ is also asymptotically exactly valid:
    \begin{equation}
        \lim_{\ell\rightarrow\infty}\mathbb{P}\big(c_{n+1}\in\ICPSet(\mathcal{T}, q_{n+1})\big)=\alpha.
    \end{equation}
\end{proposition}



\subsection{Adaptive Nonconformity Measures}\label{sec:nonconformity_measure}
While the previous section provides validity guarantees for any nonconformity measure,
standard choices such as the absolute residual (Equation \eqref{eq:absolute_residual}) lead to prediction intervals of fixed width, regardless of the uncertainty in individual test queries—thus failing to satisfy the conditionality desideratum.

To formalize this, let us denote the set of sorted nonconformity scores from the calibration set as
\begin{equation} 
\{s_{m+1}, s_{m+2}, \dots, s_{n}\} = \{s_{(1)}, s_{(2)}, \dots, s_{(\ell)}\},
\end{equation}
with $s_{(1)} \leq s_{(2)} \leq \dots \leq s_{(\ell)}$.
Then Equation~\eqref{eq:split_set_construct} can be equivalently expressed as 
\begin{align} 
\ICPSet(\mathcal{T}, &q_{n+1}) := \\
\notag&\big\{c \in [0,1] : s_{n+1} \leq s_{(\lceil \alpha(\ell+1) \rceil)}\big\}. 
\end{align}

By definition, $s_{n+1}=|M(q_{n+1})-c|$, which results in a symmetric prediction interval with absolute residual as nonconformity measure:
\begin{align} 
&\ICPSet(\mathcal{T}, q_{n+1}) := \\
\notag&\big[M(q_{n+1}) - s_{(\lceil \alpha(\ell+1) \rceil)}, M(q_{n+1}) + s_{(\lceil \alpha(\ell+1) \rceil)}\big],
\end{align}
where $M$ is the shorthand notation of a fixed UnKGE model trained on the proper training set. Since $s_{(\lceil \alpha(\ell+1) \rceil)}$ is shared across all test queries, the interval width remains constant and does not reflect query-specific uncertainty.



To address this limitation, we introduce an \emph{entropy-normalized absolute residual} as our nonconformity measure: 
\begin{equation}
\label{eq:nonconformity} 
S\big(\TrainingSet, (q,c)\big) := \Bigl|\frac{M(q) - c}{H\big(M(q)\big)}\Bigr|, 
\end{equation} 
where the normalization term $H\big(M(q)\big)$ is the entropy of the model’s prediction: 
\begin{align} 
\label{eq:entropy}
H\big(M(q)\big) := &-M(q)\log M(q) \\
\notag&- \big(1 - M(q)\big)\log\big(1 - M(q)\big). 
\end{align}

For a new query $q_{n+1}$, our method \textsc{UnKGCP} constructs the following prediction interval:
\begin{align}
\UnKGCPSet(\mathcal{T}, &q_{n+1}) := \\
&\notag\big[M(q_{n+1}) - \epsilon, M(q_{n+1}) + \epsilon\big],
\end{align}
where $\epsilon=s_{(\lceil \alpha(\ell+1) \rceil)}\cdot H\big(M(q)\big)$ is the query-specific tolerance.

This nonconformity measure scales the residual by the model’s predictive uncertainty, allowing predictions with higher entropy (i.e., lower confidence) to tolerate larger residuals.
As a result, the prediction intervals adapt to the local difficulty of each query.
Importantly, since conformal prediction ensures validity under the i.i.d. assumption regardless of the specific nonconformity measure, our approach retains its theoretical validity guarantees while producing more informative, adaptive intervals, as supported by the empirical results in Table~\ref{tab:pos_main} and Figure~\ref{fig:conditionality_nl27k}. 





\section{Experiments}\label{sec:exp}


\begin{table*}[h!]
\centering
\resizebox{0.9\textwidth}{!}{%
\begin{tabular}{@{}cc|cc|cc|cc@{}}
\toprule[1.2pt]
\multicolumn{1}{l}{} & \multicolumn{1}{l}{} & \multicolumn{2}{c}{CN15k} & \multicolumn{2}{c}{PPI5k} & \multicolumn{2}{c}{NL27k} \\ 
\multicolumn{1}{l}{} & \multicolumn{1}{l}{} & coverage & sharpness $\downarrow$ & coverage & sharpness $\downarrow$ & coverage & sharpness $\downarrow$ \\\midrule
    & FPI & 0.80 (0.000) & 0.84 (0.002) & 0.89 (0.000) & 0.67 (0.002) & {\color[HTML]{00AA00} 1.00 (0.000)} & 0.66 (0.002) \\
    & QR & 0.09 (0.005) & 0.20 (0.008) & 0.41 (0.438) & 0.40 (0.375) & {\color[HTML]{00AA00} 0.96 (0.107)} & 0.99 (0.001) \\
    & CP & {\color[HTML]{00AA00} 0.90 (0.002)} & 0.88 (0.002) & {\color[HTML]{00AA00} 0.90 (0.001)} & \textbf{0.16 (0.002)} & {\color[HTML]{00AA00} 0.90 (0.001)} & \textbf{0.27 (0.006)} \\
\multirow{-4}{*}{UKGE} & \textbf{UnKGCP} & {\color[HTML]{00AA00} 0.90 (0.001)} & \textbf{0.82 (0.003)} & {\color[HTML]{00AA00} 0.90 (0.001)} & \textbf{0.16 (0.002)} & {\color[HTML]{00AA00} 0.91 (0.003)} & 0.43 (0.018) \\\midrule
    & FPI & 0.39 (0.135) & 0.71 (0.002) & 0.89 (0.000) & 0.68 (0.001) & {\color[HTML]{00AA00} 1.00 (0.000)} & 0.76 (0.001) \\
    & QR & - & - & - & - & - & - \\
    & CP & {\color[HTML]{00AA00} 0.90 (0.002)} & 0.86 (0.003) & {\color[HTML]{00AA00} 0.90 (0.001)} & 0.21 (0.002) & {\color[HTML]{00AA00} 0.90 (0.001)} & 0.54 (0.008) \\
\multirow{-4}{*}{PASSLEAF} & \textbf{UnKGCP} & {\color[HTML]{00AA00} 0.90 (0.002)} & \textbf{0.84 (0.003)} & {\color[HTML]{00AA00} 0.90 (0.001)} & \textbf{0.20 (0.002)} & {\color[HTML]{00AA00} 0.90 (0.001)} & \textbf{0.44 (0.005)} \\\midrule
    & FPI & 0.79 (0.001) & 0.70 (0.018) & 0.89 (0.000) & 0.69 (0.006) & {\color[HTML]{00AA00} 1.00 (0.000)} & 0.67 (0.002) \\
    & QR & 0.59 (0.010) & 0.70 (0.007) & {\color[HTML]{00AA00} 0.90 (0.002)} & 0.49 (0.001) & 0.48 (0.002) & 0.86 (0.001) \\
    & CP & {\color[HTML]{00AA00} 0.90 (0.001)} & 0.86 (0.003) & {\color[HTML]{00AA00} 0.90 (0.003)} & \textbf{0.25 (0.008)} & {\color[HTML]{00AA00} 0.90 (0.002)} & 0.42 (0.006) \\
\multirow{-4}{*}{BEUrRE} & \textbf{UnKGCP} & {\color[HTML]{00AA00} 0.90 (0.001)} & \textbf{0.81 (0.003)} & {\color[HTML]{00AA00} 0.90 (0.002)} & 0.26 (0.008) & {\color[HTML]{00AA00} 0.90 (0.002)} & \textbf{0.38 (0.003)} \\ 
\bottomrule[1.2pt]
\end{tabular}%
}
\caption{Coverage and sharpness results on test triples across three datasets (CN15k, PPI5k, NL27k). We report the average over 10 trials, with standard deviation shown in parentheses. Coverage values $\geq 0.90$ are highlighted in {\color[HTML]{00AA00} green}. Among those, the method achieving the best (i.e., lowest) sharpness is \textbf{bolded}. Since \textsc{QR} is not directly applicable within the semi-supervised learning framework, no results are reported for \textsc{QR} in \textsc{PASSLEAF}.}
\label{tab:pos_main}
\end{table*}

\subsection{Experimental Settings}
\paragraph{Datasets.} 
We evaluate our method on three commonly used benchmarks: CN15k, NL27k, and PPI5k.
CN15k is a subgraph of ConceptNet \citep{DBLP:conf/aaai/SpeerCH17}, a commonsense KG.
NL27k is derived from NELL \citep{DBLP:journals/cacm/MitchellCHTYBCM18}, an automatically constructed KG from web data. 
The confidence scores in these datasets are interpreted as subjective beliefs, representing the system’s internal estimate of how likely a statement is to be true based on prior knowledge or heuristics.
PPI5k is a subset of the STRING Protein-Protein Interaction Knowledge Base \citep{DBLP:journals/nar/SzklarczykMC0WS17}, where scores correspond to statistical probabilities derived from experimental evidence.
Dataset statistics are summarized in Table~\ref{tab:data_statistics}, with additional details provided in Appendix~\ref{app:dataset}.


\begin{table}[h!]
\centering
\resizebox{0.49\textwidth}{!}{%
\begin{tabular}{c|cc|cc|cc}
\toprule[1.2pt]
 & \multicolumn{2}{c}{CN15k} & \multicolumn{2}{c}{PPI5k} & \multicolumn{2}{c}{NL27k}\\
 & MSE $\downarrow$ & MAE $\downarrow$ & MSE $\downarrow$ & MAE $\downarrow$ & MSE $\downarrow$ & MAE $\downarrow$
\\\midrule
UKGE & 0.24 & 0.41 & 0.01 & 0.04 & 0.05 & 0.11 \\
PASSLEAF & 0.24 & 0.41 & 0.01 & 0.03 & 0.06 & 0.11 \\
BEUrRE & 0.12 & 0.28 & 0.01 & 0.06 & 0.03 & 0.12 \\
\bottomrule[1.2pt]
\end{tabular}%
}
\caption{Mean squared error (MSE) and mean absolute error (MAE) of the UnKGE models. We report the mean over 10 trials; the standard deviation is negligible.}
\label{tab:performance}
\end{table}

\paragraph{UnKGE Backbones.}
We base our experiments on three representative UnKGE methods: UKGE \citep{chen2019embedding}, PASSLEAF \citep{chen2021passleaf}, and BEUrRE \citep{chen2021probabilistic}. Detailed descriptions of each method are provided in Appendix~\ref{app:backbone}.

\paragraph{Confidence Predictors.}
We compare our proposed method against three established baseline techniques for constructing prediction intervals: (1) Fisher Prediction Intervals (FPI) \cite{fisher1935fiducial}, (2) Quantile Regression (QR) \cite{koenker1978regression}, and (3) Conformal Prediction (CP) \cite{vovk2005algorithmic} with absolute residuals as the nonconformity measure. Additional details are provided in Appendix~\ref{app:confidence_predictor}.

\paragraph{Evaluation Metrics.}
\label{sec:metrics}
We evaluate prediction intervals using two standard metrics: \emph{Coverage} and \emph{Sharpness}.
Given a test set $\TestSet = \{(q_i,c_i)\}_{i=n+1}^{N}$ of size $k = N - n$, these metrics are formally defined as follows:

\textbf{Coverage} measures the fraction of test queries for which the ground-truth confidence score $c_i$ is covered by the prediction interval:
\begin{equation}
\text{Coverage} = \frac{1}{k}\sum_{i=n+1}^{N}\mathbbm{1}\big[\,c_i \in \Gamma^\alpha(\mathcal{T}, q_i)\,\big],
\end{equation}
and serves as an empirical estimate of the theoretical coverage probability in Equation \eqref{eq:coverage}.

\textbf{Sharpness} quantifies the average length of the prediction intervals. Let $\Gamma^\alpha(\mathcal{T}, q_i)=[l_i, u_i]$, where $l_i$ and $u_i$ are the lower and upper bounds. Then, 
\begin{equation}
\text{Sharpness} = \frac{1}{k}\sum_{i=n+1}^{N}(u_i-l_i).
\end{equation}

An effective confidence predictor should achieve coverage at least equal to the target confidence level while maintaining the smallest possible sharpness.



\subsection{Analysis of Empirical Validity and Efficiency}

\begin{figure*}[h!]
    \centering
    \includegraphics[width=\textwidth]{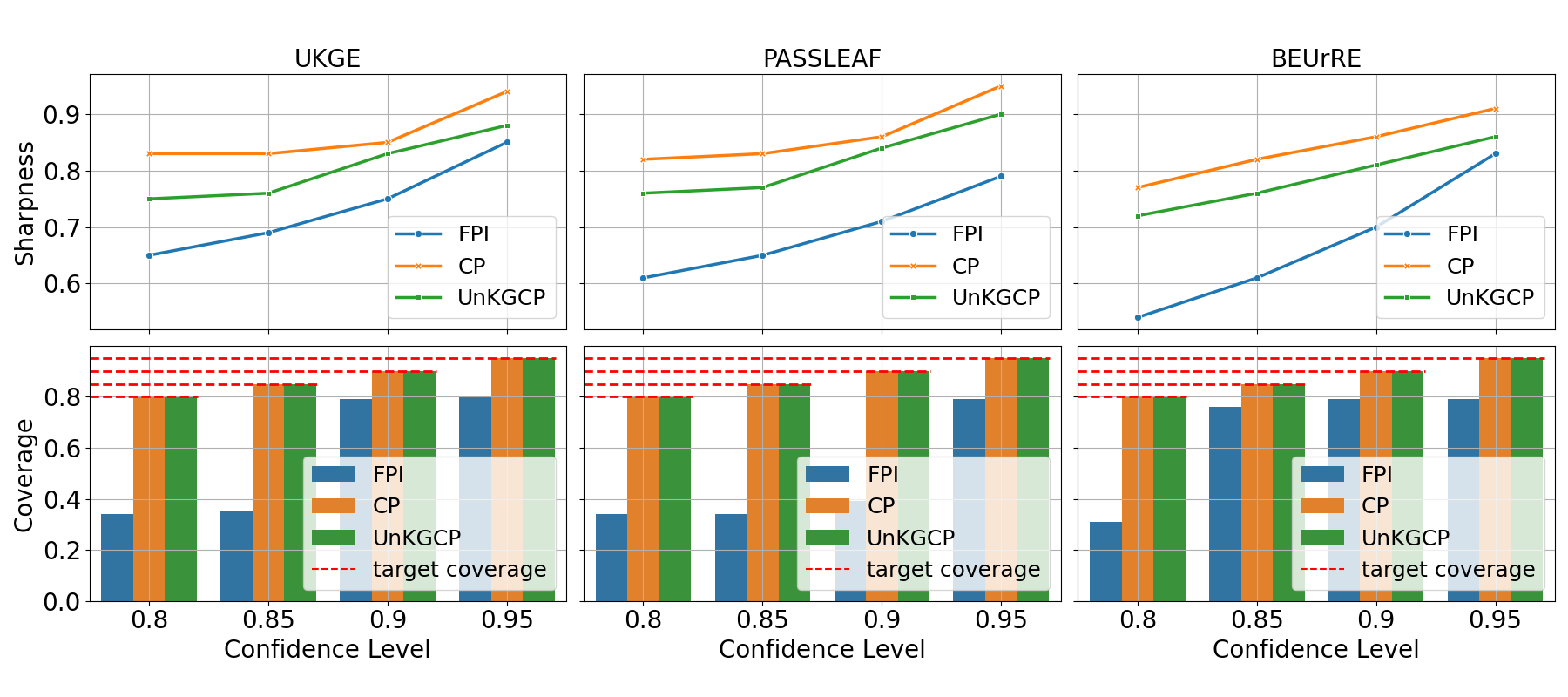}
    \caption{Effect of the confidence level $\alpha$ on the sharpness (top) and coverage (bottom) for test triples on CN15k. Each curve represents one predictor. Red dashed lines indicate the desired coverage levels. Additional results can be found in Figures~\ref{fig:confidence_level_NL27k_pos}--\ref{fig:confidence_level_PPI5k_neg} in Appendix~\ref{app: complete_results}.}
    \label{fig:confidence_level_CN15k_pos}
\end{figure*}

We evaluate the empirical performance of confidence predictors in terms of \textit{validity} and \textit{efficiency}, using coverage and sharpness as defined in Section~\ref{sec:metrics}. 
Table~\ref{tab:pos_main} summarizes the results on test triples across the three benchmark datasets and the UnKGE backbones at a $90\%$ confidence level.

\textbf{The validity guarantees in Proposition~\ref{th:validity} are empirically supported}, as both conformal predictors (\textsc{CP} and \textsc{UnKGCP}) achieve coverage probabilities closely matching the target confidence across all dataset-backbone configurations.
In contrast, other baseline confidence predictors (\textsc{Fisher} and \textsc{QR}) often fail to achieve the target coverage, especially on CN15k and PPI5k. This is likely because \textsc{Fisher} assumes normally distributed residuals—a condition rarely met in UnKGs—and \textsc{QR} relies on a well-specified conditional quantile model, which is challenging given the limited expressiveness of current UnKGE methods.

In terms of efficiency, \textsc{CP} and \textsc{UnKGCP} consistently produce sharper prediction intervals on PPI5k and NL27k. While \textsc{Fisher} and \textsc{QR} yield narrower intervals on CN15k, this comes at the cost of systematic undercoverage. \textbf{Notably, our proposed \textsc{UnKGCP} outperforms \textsc{CP} in 7 out of 9 configurations by generating sharper intervals}. 
Moreover, the average interval lengths from valid confidence predictors (\textsc{CP} and \textsc{UnKGCP}) correlate with UnKGE model performance across datasets (Table~\ref{tab:performance}): wider intervals are associated with higher uncertainty and lower model performance, demonstrating that conformal predictors effectively capture model-level uncertainty through interval length.

In summary, \textsc{UnKGCP} achieves the best overall performance by simultaneously satisfying the validity criterion and generating reasonably sharp prediction intervals. 
Figure~\ref{fig:confidence_level_CN15k_pos} further illustrates the performance of all confidence predictors across multiple confidence levels ranging from 80\% to 95\% in increments of 5\%. The conclusion remains consistent: \textbf{\textsc{UnKGCP} maintains superior performance across all confidence levels}. Notably, the length of the prediction intervals increases with higher confidence levels, aligning with the monotonicity property described in Equation~\eqref{eq:nonconformity}.

\begin{figure*}[t]
    \centering
    \includegraphics[width=0.9\textwidth]{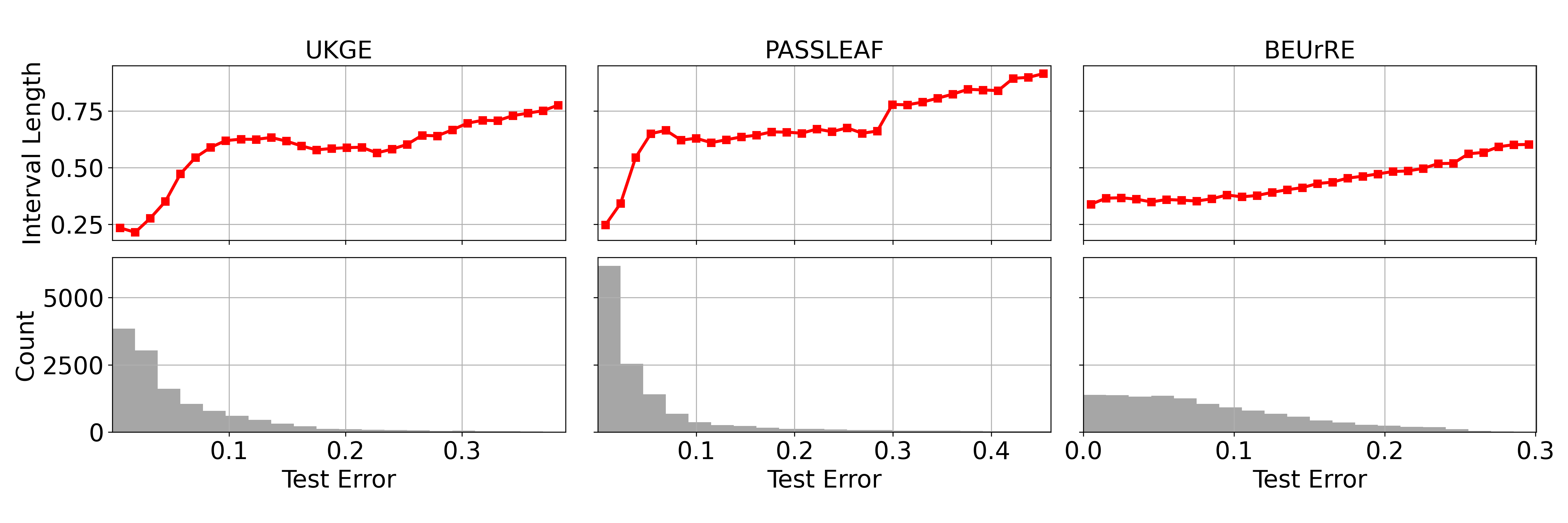}
    \caption{Conditionality analysis on NL27k. Each column corresponds to a different backbone model (\textsc{BEUrRE}, \textsc{UKGE}, \textsc{PASSLEAF}). Top: test instances are grouped into 30 bins based on prediction error, and the mean prediction interval length is computed per bin. Only intervals that cover the ground truth are included, as non-covering intervals are not expected to reflect query difficulty. Bottom: histogram of test errors is shown to illustrate their distribution. The complete results are provided in Figures \ref{fig:conditionality_cn15k}--\ref{fig:conditionality_ppi5k_neg} in Appendix \ref{app: complete_results}.}
    \label{fig:conditionality_nl27k}
\end{figure*}

\subsection{Analysis of Conditionality}
As discussed in Section \ref{sec:nonconformity_measure}, \textsc{CP} produces fixed-length prediction intervals and thus fails to satisfy the conditionality desideratum. 
In this section, we show that \textsc{UnKGCP} not only produces valid and sharper prediction intervals but also \textbf{outperforms \textsc{CP} and other baselines by adapting interval lengths to query difficulty}.


Following \citet{zhu2025conformalized, Angelopoulos2021RAPS}, we use the \emph{absolute prediction error} as a proxy for instance-level difficulty, with larger errors indicating harder queries. 
Figure~\ref{fig:conditionality_nl27k} shows that, across all models, average interval length increases with error—demonstrating that prediction intervals adapt to instance difficulty and thus satisfy the conditionality criterion. \textbf{This implies that uncertainty can be reliably inferred from the interval length produced by \textsc{UnKGCP}}.

\subsection{Impact of Calibration Set Size}
While conformal prediction offers validity guarantees under i.i.d~\citep{vovk2005algorithmic, lei2018distribution}, small calibration sets may yield high-variance estimates of the nonconformity threshold. This can lead to unstable prediction intervals that either under-cover or become unnecessarily wide in practice. In this section, we study how the size of the calibration set influences the performance of \textsc{UnKGCP} in terms of coverage and sharpness.

\begin{figure}[t]
    \centering
    \includegraphics[width=0.9\linewidth]{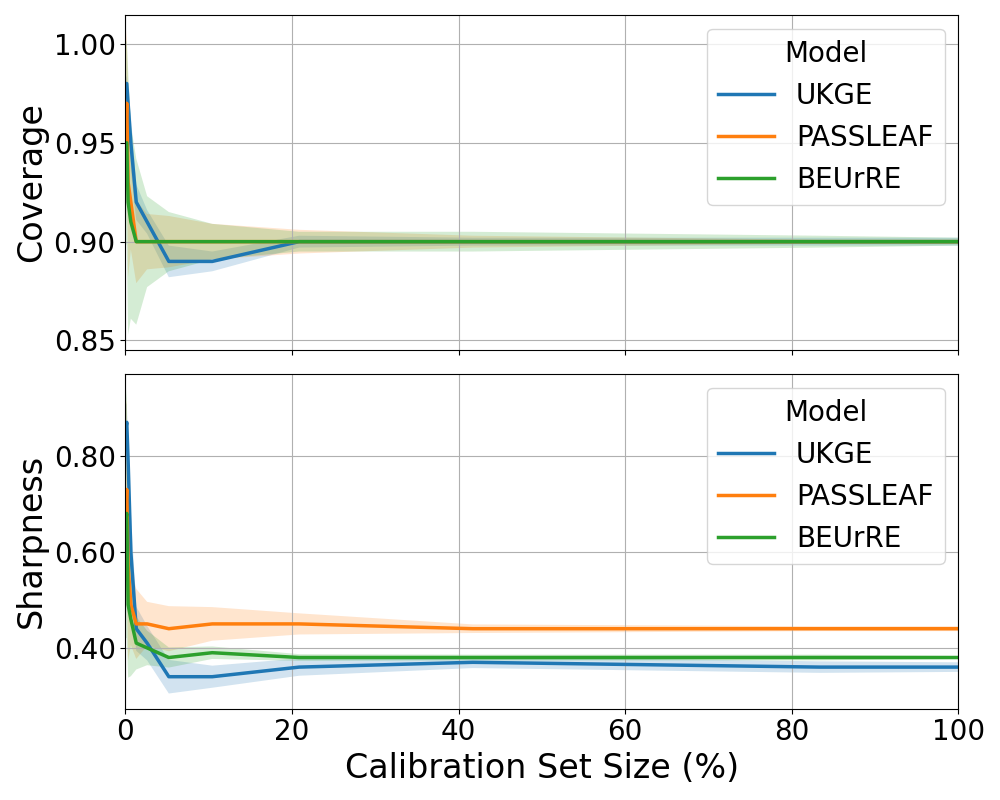}
    \caption{Effect of calibration set size on coverage and sharpness on NL27k. The top panel reports coverage and the bottom panel reports sharpness. In both plots, the lines represent mean values across 10 runs, and the shaded areas indicate standard deviation. The complete results are provided in Figures \ref{fig:calib_size_analysis_cn15k_pos}--\ref{fig:calib_size_analysis_ppi5k_neg} in Appendix \ref{app: complete_results}.}
    \label{fig:calib_size_analysis_nl27k_pos}
\end{figure}

We randomly sample increasingly larger subsets of the calibration set—starting from 10 triples and doubling the size each time (i.e., 10, 20, 40, ...)—until the full set is used. For each subset size, we repeat the sampling 10 times and report the mean and standard deviation of coverage and sharpness.
Figure~\ref{fig:calib_size_analysis_nl27k_pos} summarizes the results for three UnKGE-based models on NL27k.

When the calibration set is small (e.g., less than 5–10\% of the data), we observe significant variability in both coverage and sharpness.
This is caused by unreliable quantile estimates, as limited calibration data can produce a biased distribution of nonconformity scores.
As a result, the prediction intervals either under-cover or become overly conservative.
As the calibration size increases, both metrics stabilize rapidly. \textbf{Notably, \textsc{UnKGCP} demonstrates strong sample efficiency: using only about 20\% of the calibration data is sufficient to achieve reliable and stable performance across all models.}


\section{Sensitivity to Distribution Shift}
\begin{table*}[t!]
\centering
\resizebox{0.9\textwidth}{!}{%
\begin{tabular}{@{}cc|cc|cc|cc@{}}
\toprule[1.2pt]
\multicolumn{1}{l}{} & \multicolumn{1}{l}{} & \multicolumn{2}{c}{CN15k} & \multicolumn{2}{c}{PPI5k} & \multicolumn{2}{c}{NL27k} \\ 
\multicolumn{1}{l}{} & \multicolumn{1}{l}{} & coverage & sharpness & coverage & sharpness & coverage & sharpness \\\midrule
    & FPI & {\color[HTML]{00AA00} 1.00 (0.000)} & \textbf{0.22 (0.002)} & {\color[HTML]{00AA00} 1.00 (0.000)} & \textbf{0.11 (0.001)} & {\color[HTML]{00AA00} 1.00 (0.000)} & \textbf{0.15 (0.002)} \\
    & QR & 0.00 (0.000) & 0.20 (0.008) & 0.17 (0.312) & 0.40 (0.376) & 0.70 (0.481) & 1.00 (0.001) \\
    & CP & 0.82 (0.003) & 0.26 (0.002) & 0.78 (0.003) & \textbf{0.05 (0.002)} & 0.79 (0.007) & 0.12 (0.004) \\
\multirow{-4}{*}{UKGE} & \textbf{UnKGCP} & 0.82 (0.003) & \textbf{0.26 (0.002)} & 0.78 (0.003) & \textbf{0.05 (0.001)} & 0.79 (0.007) & \textbf{0.08 (0.003)} \\\midrule
    & FPI & {\color[HTML]{00AA00} 1.00 (0.000)} & \textbf{0.11 (0.004)} & {\color[HTML]{00AA00} 1.00 (0.000)} & \textbf{0.10 (0.005)} & {\color[HTML]{00AA00} 1.00 (0.000)} & \textbf{0.12 (0.007)} \\
    & QR & - & - & - & - & - & - \\
    & CP & 0.75 (0.003) & 0.10 (0.003) & 0.71 (0.006) & \textbf{0.04 (0.001)} & 0.70 (0.009) & \textbf{0.06 (0.002)} \\
\multirow{-4}{*}{PASSLEAF} & \textbf{UnKGCP} & 0.75 (0.003) & \textbf{0.10 (0.003)} & 0.71 (0.008) & 0.06 (0.001) & 0.70 (0.009) & 0.07 (0.002) \\\midrule
    & FPI & {\color[HTML]{00AA00} 1.00 (0.000)} & \textbf{0.29 (0.008)} & {\color[HTML]{00AA00} 1.00 (0.000)} & \textbf{0.09 (0.004)} & {\color[HTML]{00AA00} 1.00 (0.000)} & 0.12 (0.005) \\
    & QR & 0.58 (0.326) & 0.58 (0.160) & 0.46 (0.334) & 0.09 (0.001) & 0.43 (0.074) & 0.06 (0.001) \\
    & CP & 0.72 (0.006) & 0.41 (0.014) & 0.75 (0.018) & \textbf{0.01 (0.006)} & {\color[HTML]{00AA00} 0.91 (0.001)} & 0.05 (0.006) \\
\multirow{-4}{*}{BEUrRE} & \textbf{UnKGCP} & 0.72 (0.006) & \textbf{0.40 (0.014)} & 0.75 (0.020) & 0.04 (0.006) & {\color[HTML]{00AA00} 0.91 (0.002)} & \textbf{0.02 (0.002)} \\ 
\bottomrule[1.2pt]
\end{tabular}%
}
\caption{Coverage and sharpness results on \textbf{negative} test triples across three datasets (CN15k, PPI5k, NL27k). We report the average over 10 trials, with standard deviation shown in parentheses. Coverage values $\geq 0.90$ are highlighted in {\color[HTML]{00AA00} green}. Among those, the method achieving the best (i.e., lowest) sharpness is \textbf{bolded}. Since \textsc{QR} is not directly applicable within the semi-supervised learning framework, no results are reported for \textsc{QR} in \textsc{PASSLEAF}.}
\label{tab:neg_main}
\end{table*}


A key assumption for the validity guarantee in Proposition~\ref{prop:main} is that data points are i.i.d. This assumption is easily violated in the negative triple setting, where negatives are synthetically generated by corrupting positives \cite{chen2019embedding}, leading to distribution shifts across training, calibration, and test sets. We therefore evaluate all methods under this setting.

Table~\ref{tab:neg_main} shows that performance on negatives is markedly less stable than on positives. \textsc{FPI} attains 100\% coverage with reasonably sharp intervals, likely because nonconformity scores for negatives concentrate near 0, making its Gaussian assumption well-suited here. Nonetheless, \textbf{\textsc{UnKGCP} is more practical and informative in real-world use}. Despite the loss of formal validity under distribution shift, it maintains strong empirical coverage—around 0.8 with UKGE and above 0.7 in most other settings, reaching 0.91 in BOX–NL27k. Crucially, unlike the fixed intervals of \textsc{FPI}, \textsc{UnKGCP} produces query-specific intervals that adapt to prediction difficulty, as illustrated in Figures~\ref{fig:conditionality_cn15k}--\ref{fig:conditionality_ppi5k_neg}.

Another interesting observation, consistent with findings by \citet{kaur2022idecode}, is that our method can be effectively used to detect significant distribution shifts—specifically, in cases where there is a substantial gap between the empirical coverage and the target confidence level.

\section{Discussion and Conclusion}
We presented \textsc{UnKGCP}, a model-agnostic uncertainty quantification framework for UnKGE models that constructs prediction intervals with formal statistical guarantees (Proposition~\ref{prop:main}).
Experiments across multiple UnKGE models and benchmarks show that \textsc{UnKGCP} produces valid, sharp, and query-adaptive intervals, where interval length reliably reflects predictive uncertainty. Additionally, \textsc{UnKGCP} is sample-efficient, achieving stable performance with only a small calibration set.

Importantly, our uncertainty estimates also offer insights that standard metrics (e.g., mean squared error) fail to capture. For instance, although all UnKGE models seem to achieve reasonably low errors in Table \ref{tab:performance} on CN15k, \textsc{UnKGCP} reveals average interval length exceeding 0.8—indicating that the predictions are rather random. This suggests limitations in either the dataset quality or model expressiveness, highlighting the critical role of uncertainty quantification in evaluating model reliability beyond point-based accuracy metrics.

\section{Limitations}
A limitation of our current method is the assumption that the input graph contains triples annotated with single-valued confidence scores. In practice, however, confidence may be expressed as intervals, or predicted intervals may be added back into the graph. In such cases, the model must be extended to handle interval-valued inputs. Specifically, each input interval can be represented by two components: its mean and its length. The UnKGE model would then be trained to predict both components. During the calibration step of conformal prediction, rather than analyzing the distribution of scalar scores, we would analyze the joint distribution of predicted means and lengths. Separate quantile thresholds would be computed for each, and two conformal intervals—one for the mean and one for the length—would be constructed and subsequently combined to form the final interval, preserving statistical coverage guarantees.

\newpage

\section{Acknowledgments}
The authors thank the International Max Planck Research School for Intelligent Systems (IMPRS-IS) for supporting Yuqicheng Zhu, Jingcheng Wu and Hongkuan Zhou. The work was partially supported by EU Projects Graph Massivizer (GA 101093202), enRichMyData (GA 101070284), SMARTY (GA 101140087), the EPSRC project OntoEm (EP/Y017706/1) and the Deutsche Forschungsgemeinschaft (DFG, German Research Foundation) - SFB 1574 - Project number 471687386. The authors also gratefully acknowledge the computing time provided on the high-performance computer HoreKa by the National High-Performance Computing Center at KIT (NHR@KIT). This center is jointly supported by the Federal Ministry of Education and Research and the Ministry of Science, Research and the Arts of Baden-Württemberg, as part of the National High-Performance Computing (NHR) joint funding program (\url{https://www.nhr-verein.de/en/our-partners}). HoreKa is partly funded by the German Research Foundation (DFG).

\newpage
\bibliography{anthology}

\newpage
\appendix
\onecolumn
\section{Proof}\label{app:proof}
\setcounter{proposition}{0}
\begin{proposition}
    Assume weighted triples in $\mathcal{T}$ and the test weighted triple $(q_{n+1}, c_{n+1})$ are i.i.d. For any confidence level $\alpha\in[0,1]$ and any nonconformity measure $S$, the ICP-based conformal predictor $\ICPSet$ is conservatively valid:
    \begin{equation}
        \mathbb{P}\big(c_{n+1}\in\ICPSet(\mathcal{T}, q_{n+1})\big)\geq\alpha.
    \end{equation}
    Furthermore, if $\{s_i\}_{i=m+1}^n$ contains no ties, $\ICPSet$ is also asymptotically exactly valid:
    \begin{equation}
        \lim_{\ell\rightarrow\infty}\mathbb{P}\big(c_{n+1}\in\ICPSet(\mathcal{T}, q_{n+1})\big)=\alpha.
    \end{equation}
\end{proposition}

\begin{proof}[Proof of the lower bound]
    Recall that the set of weighted triples $\mathcal{T}$ is partitioned into a proper training set $\TrainingSet=\{(q_i, c_i)\}_{i=1}^{m}$ and a calibration set $\CalibrationSet=\{(q_i, c_i)\}_{i=m+1}^{n}$ of size $\ell=n-m$.
    
    By assuming all weighted triples
    \begin{equation}
        (q_{m+1}, c_{m+1}), \dots, (q_{n+1}, c_{n+1})
    \end{equation}
    are i.i.d, we know that their order is a uniform random permutation of the indices $m+1, \dots, n+1$.
    Hence the corresponding nonconformity scores $\{s_{m+1}, \dots, s_{n+1}\}$ are exchangeable: every one of the $(\ell+1)!$ permutations of these scores is equally likely \cite[Section 3]{papadopoulos2002inductive}, \cite[Chapter 4.2.2]{vovk2005algorithmic}. Formally, for any permutation $\pi:\{m+1, \dots, n+1\}\rightarrow\{m+1, \dots, n+1\}$,
    \begin{equation}\label{eq:exchangeability}
        (s_{m+1}, \dots, s_{n+1})\overset{d}{=}(s_{\pi(m+1)}, \dots, s_{\pi(n+1)}),
    \end{equation}
    where $\overset{d}{=}$ denotes equality in distribution.
    
    The ICP-based prediction interval at confidence level $\alpha$ includes a candidate score if and only if $s_{n+1}$ is among the $\lceil \alpha(\ell+1)\rceil$ smallest $s_i$:
    \begin{equation}
        \frac{|\{i=m+1,\dots,n+1:s_i\geq s_{n+1}\}|}{\ell+1}>1-\alpha.
    \end{equation}
    
    Due to the exchangeability in Equation \eqref{eq:exchangeability}, each of the $\ell+1$ positions that $s_{n+1}$ could occupy among the scores $\{s_{m+1}, \dots, s_{n+1}\}$ is equally likely. Thus, the probability that the prediction interval covers the ground truth scores equals
    \begin{equation}
        \mathbb{P}\big(c_{n+1}\in\ICPSet(\mathcal{T}, q_{n+1})\big)=\frac{\lceil \alpha(\ell+1)\rceil}{\ell+1}\geq\alpha,
    \end{equation}
    which establishes conservative coverage at level $\alpha$.
\end{proof}

\begin{proof}[Proof of the upper bound]
    We prove the upper bound based on \citet[Appendix A.1]{lei2018distribution}.
    By assuming no ties in the set of nonconformity scores in $\{s_{m+1}, \dots, s_{n+1}\}$, we know that the  nonconformity scores in $\{s_{m+1}, \dots, s_{n+1}\}$ are all distinct with probability one.
    Define $\alpha'=\alpha+1/(\ell+1)$.
    Consider now the complementary set $\ComICPSet$:
    \begin{equation}
    \ComICPSet(\mathcal{T}, q_{n+1}):=\bigg\{c\in[0,1]:\frac{|i=m+1,\dots,n+1:s_i\geq s_{n+1}|}{\ell+1}\leq1-\alpha'\bigg\},
    \end{equation}
    where 
    \begin{align}
        & s_i:=S\big(\TrainingSet, (q_i, c_i)\big), i=m+1,\dots, n,\\
        \notag& s_{n+1}:=S\big(\TrainingSet, (q_{n+1}, c)\big).
    \end{align}
    Due to the i.i.d. assumption and hence exchangeability of the nonconformity scores $\{s_{m+1}, \dots, s_{n+1}\}$, the rank of $s_{n+1}$ among these $\ell + 1$ scores is uniformly distributed. Therefore, for any fixed $c \in [0,1]$,
    \begin{equation}
        \mathbb{P}\big(c_{n+1}\in\ComICPSet(\mathcal{T}, q_{n+1})\big)=\frac{\lceil (1-\alpha')(\ell+1)\rceil}{\ell+1}\geq1-\alpha'=1-\alpha-\frac{1}{\ell+1}
    \end{equation}
    Moreover, since we assumed no ties, the sets $\ICPSet(\mathcal{T}, q_{n+1})$ and $\ComICPSet(\mathcal{T}, q_{n+1})$ are disjoint:
    \begin{equation}
        \ICPSet(\mathcal{T}, q_{n+1})\cap\ComICPSet(\mathcal{T}, q_{n+1})=\emptyset
    \end{equation}
    Thus,
    \begin{align*}
        &\mathbb{P}\big(c_{n+1}\in\ICPSet(\mathcal{T}, q_{n+1})\big)+\mathbb{P}\big(c_{n+1}\in\ComICPSet(\mathcal{T}, q_{n+1})\big)\leq 1\\
        \Rightarrow&\mathbb{P}\big(c_{n+1}\in\ICPSet(\mathcal{T}, q_{n+1})\big)\leq 1-\mathbb{P}\big(c_{n+1}\in\ComICPSet(\mathcal{T}, q_{n+1})\big)\\
        \Rightarrow&\mathbb{P}\big(c_{n+1}\in\ICPSet(\mathcal{T}, q_{n+1})\big)\leq \alpha+\frac{1}{\ell+1}
    \end{align*}
    Combine this upper bound with the lower bound proved earlier; together they give
    \begin{equation}
        \alpha\leq\mathbb{P}\big(c_{n+1}\in\ICPSet(\mathcal{T}, q_{n+1})\big)\leq\alpha+\frac{1}{\ell+1}
    \end{equation}
    Hence $\mathbb{P}\big(c_{n+1}\in\ICPSet(\mathcal{T}, q_{n+1})\big)\xrightarrow[\ell\to\infty]{}\alpha$ establishing \emph{exact} asymptotic validity when ties occur with probability 0.
\end{proof}

\twocolumn

\section{Details of Experimental Settings}
\subsection{Details of Datasets}\label{app:dataset}
We provide further details on the three benchmark datasets used in our experiments: CN15k, NL27k, and PPI5k.

\textbf{CN15k} is derived from ConceptNet \citep{DBLP:conf/aaai/SpeerCH17}, a multilingual commonsense knowledge graph. Each assertion has a confidence score between 0.1 and 22, with 99.6\% of scores below or equal to 3.0. Following \citet{chen2019embedding}, we cap the scores at 3.0 and then apply $log$ by min-max normalization to map the values into the range $[0.5, 1.0]$.

\textbf{NL27k} is built from NELL \citep{DBLP:journals/cacm/MitchellCHTYBCM18}, a large-scale English knowledge base constructed via semi-automatic extraction. Confidence scores are assigned based on iterative self-training and rule-based extraction pipelines. These scores, originally in $[0.1, 0.9]$, are min-max normalized to $[0.1, 1.0]$.

\textbf{PPI5k} is derived from STRING \citep{DBLP:journals/nar/SzklarczykMC0WS17}, a biological database of protein-protein interactions. Each triple represents an interaction between two proteins, annotated with a confidence score between 0 and 1. These confidence scores can be interpreted as probabilities. For example, a score of 0.5 indicates that about half of the predicted interactions may be false positives. Higher scores suggest higher probability for a true biological interaction.

\paragraph{Data Splits.}
Following \citet{chen2019embedding}, all datasets are partitioned into 85\% for training, 7\% for calibration, and 8\% for testing. The data statistics and splits are shown in Table~\ref{tab:data_statistics}. These are used consistently across all models and experiments.

\begin{table}[h!]
\centering
\resizebox{0.49\textwidth}{!}{%
\begin{tabular}{cccc}
\toprule
Dataset & \#Entity & \#Predicate & \#Training/Calibration/Test Facts
\\\midrule
CN15k & 15,000 & 36 &  204,984/16,881/19,293\\
NL27k & 27,221 & 404 &  149,100/12,278/14,034\\
PPI5k & 4,999 & 7 &  230,929/19,017/21,720\\
\bottomrule
\end{tabular}%
}
\caption{Dataset statistics used in our experiments.}
\label{tab:data_statistics}
\end{table}

\subsection{Details of UnKGE Backbones}\label{app:backbone}
\textbf{UKGE} \cite{chen2019embedding} extends KGE methods by explicitly modeling uncertainty through confidence scores with each triple. It adapts the scoring function and loss function to predict continuous values in $[0, 1]$ that reflect the plausibility of triples. Given a triple $\triple$, and following the DistMult model \citep{DBLP:journals/corr/YangYHGD14a}, the UKGE score function is defined as:
\begin{equation}
f\left((\mathbf{h} \circ \mathbf{t})^\top \mathbf{r} \right)
\end{equation}
where $\mathbf{h}$, $\mathbf{r}$, and $\mathbf{t}$ denote the embeddings of the head entity, relation, and tail entity, respectively and $f:\mathbb{R}\rightarrow[0,1]$ is a normalization function that maps the raw score to a confidence score in $[0, 1]$.

UKGE offers two variants, UKGE$_\text{logi}$ and UKGE$_\text{rect}$, which differ in how they map raw triple scores to confidence scores. We primarily focus on UKGE$_\text{logi}$, which applies a learnable logistic function to map the raw triple score to a confidence score. Specifically, given a triple $\triple$ with embeddings $\mathbf{h}$, $\mathbf{r}$, and $\mathbf{t}$, the score is computed as:
\begin{equation}
f_{\text{logi}}(h, r, t) = \frac{1}{1 + \exp\left(- \left(w (\mathbf{h} \circ \mathbf{t})^\top \mathbf{r} + b\right)\right)},
\end{equation}
where $w$ and $b$ are learnable scalar parameters of the logistic function. 

An alternative variant, UKGE$_\text{rect}$, adopts a rectified and bounded linear transformation:
\begin{align}
&f_{\text{rect}}(h, r, t) = \\
&\notag\min\left( \max\big(w (\mathbf{h} \circ \mathbf{t})^\top \mathbf{r} + b, 0\big), 1 \right).
\end{align}

Probabilistic Soft Logic (PSL) \cite{Kimmig2012ASI} is used to estimate confidence scores $c$ for unseen weighted triples, resulting in an extended weighted triple set $\PSLTripleSet$ that augments the original training dataset. We define the augmented training set as $\mathcal{T}^+=\mathcal{T}\cup\PSLTripleSet$ and let $\mathcal{T}^-$ denote the set f negative weighted triples. The loss is then defined as:
\begin{equation}
L = \sum_{(q, c) \in \mathcal{T}^+} \left( \model(q) - c \right)^2 + \alpha \sum_{(q, c) \in \NegTripleSet}\model(q)^2,
\label{eq:ukge_loss}
\end{equation}
Note that $c$ is $0$ for weighted triples in $\NegTripleSet$, and $\alpha \in \mathbb{R}^+$ is a hyperparameter controlling the penalty on unobserved triples in $\NegTripleSet$.

\textbf{PASSLEAF}~\cite{chen2021passleaf} extends the UKGE framework by incorporating a semi-supervised learning strategy that leverages pseudo-labeled triples to better utilize uncertain information. Its training objective consists of three components: a supervised loss over observed positive triples, a loss over generated negative triples, and a semi-supervised loss over pseudo-labeled triples.

The supervised loss $L_{\text{pos}}$ minimizes the mean squared error between the model’s predicted confidence scores and the ground-truth scores $c$ for each weighted triple $(q, c) \in \PosTripleSet$:
\begin{equation}
L_{\text{pos}} = \sum_{(q, c)\in\PosTripleSet} \left| \model(q) - c \right|^2.
\end{equation}

The negative sample loss $L_{\text{neg}}$ encourages the model to assign low confidence scores to generated negative triples $(q, c) \in \NegTripleSet$:
\begin{equation}
L_{\text{neg}} = \sum_{(q, c) \in \NegTripleSet} \left| \model(q) \right|^2.
\end{equation}

PASSLEAF introduces an additional semi-supervised loss $L_{\text{semi}}$ over a pseudo-labeled set $\SemiTripleSet$, where each query triple is assigned a confidence score based on the model’s own predictions from a prior training stage.
\begin{equation} 
L_{\text{semi}} = \sum_{(q, c) \in \SemiTripleSet} \left| \model(q) - c \right|^2.
\end{equation}

The overall objective combines all components, with the semi-supervised and negative losses normalized by the total number of generated triples:
\begin{equation}
L = L_{\text{pos}} + \frac{1}{|\SemiTripleSet \cup \NegTripleSet|} \left( L_{\text{semi}} + L_{\text{neg}} \right).
\end{equation}

\textbf{BEUrRE}~\cite{chen2021probabilistic} models entities and relations as probabilistic boxes in a latent space. This approach supports detailed modeling of uncertainty at both the fact and entity levels. 
Given a triple $q = \triple$, the model defines the confidence score $\model(q)$ as an approximate conditional probability:
\begin{equation}
\model(q) = \frac{ \mathbb{E} \left[ \text{Vol}\left( H_r(Box_h) \cap T_r(Box_t) \right) \right] }{ \mathbb{E} \left[ \text{Vol}\left( T_r(Box_t) \right) \right] },
\end{equation}
where $Box_h$ and $Box_t$ denote the probabilistic boxes corresponding to entities $h$ and $t$, respectively. The functions $H_r$ and $T_r$ are relation-specific transformations, typically implemented as affine mappings. $\text{Vol}(\cdot)$ denotes the volume of a box, and $\mathbb{E}[\cdot]$ represents the expectation taken over the stochastic parameters of the boxes. BEUrRE is trained using the same loss function as defined in Equation~\eqref{eq:ukge_loss}. 

\subsection{Details of Confidence Predictor}\label{app:confidence_predictor}
\subparagraph{Fisher Prediction Intervals (FPI)} \cite{fisher1935fiducial} provide a classical statistical approach to quantifying uncertainty in predicted confidence scores. FPI relies on assumptions: (i) the residuals (i.e. additive noise of the regression model) are approximately normally distributed, (ii) the calibration examples are independent and identically distributed (i.i.d.), and (iii) the prediction variance is constant across instances (homoscedasticity). Under these assumptions, FPI offers a parametric interval estimation framework using the $t$-distribution. Given a \emph{calibration set} $\CalibrationSet=\{tr_i\}_{i=m+1}^{n}$ of size $\ell = n - m$, consisting of predicted confidence scores $\{c_{m+1}, \dots, c_n\}$, we compute the sample mean $\bar{c}_{\text{cal}}$ and unbiased sample variance $s^2_{\text{cal}}$ as:
\begin{equation}
\bar{c}_{\text{cal}} = \frac{1}{\ell} \sum_{i=m+1}^{n} c_i,
\end{equation}
\begin{equation}
s^2_{\text{cal}} = \frac{1}{\ell - 1} \sum_{i=m+1}^{n} (c_i - \bar{c}_{\text{cal}})^2.
\end{equation}

The FPI-based prediction interval for a new query $q_{n+1}$ is given by:
\begin{equation}
\begin{split}
\FisherSet(\mathcal{T}, q_{n+1}) := \Bigg[\bar{c}_{\text{cal}} - t_{\ell-1}^{(1-\alpha)/2} \cdot s_{\text{cal}} 
\sqrt{\frac{\ell}{\ell-1}}, \\
\bar{c}_{\text{cal}} + t_{\ell-1}^{(1-\alpha)/2} \cdot s_{\text{cal}} \sqrt{\frac{\ell}{\ell-1}}\Bigg],
\end{split}
\end{equation}
where $t_{\ell-1}^{(1-\alpha)/2}$ is the $(1-\alpha)/2$ quantile of the $t$-distribution with $\ell - 1$ degrees of freedom. For instance, setting $\alpha = 0.9$ yields a prediction interval with 90\% confidence.

\subparagraph{Quantile Regression (QR)} \cite{koenker1978regression} provides a flexible approach to modeling the conditional distribution of predicted confidence scores without assuming any specific parametric form. Given the \emph{proper training set} $\TrainingSet=\{tr_i\}_{i=1}^{m}$, we train two separate quantile regressors to construct prediction intervals: a lower quantile model at $\tau_{\text{lower}} = (1 - \alpha)/2$ and an upper quantile model at $\tau_{\text{upper}} = 1 - (1 - \alpha)/2$, where $\alpha$ denotes the desired confidence level.

Each quantile model $\model$ is optimized by minimizing the pinball loss function:
\begin{align} 
L = \sum_{(q,c)\in\TrainingSet} &\tau \cdot \max\big(c - \model(q), 0\big)\\
&\notag +(1-\tau)\cdot\max\big(\model(q)-c, 0\big),
\end{align}
The resulting prediction interval for a new query $q_{n+1}$ is obtained by evaluating the two trained models with confidence level $\alpha$:
\begin{align}
&\QuantileSet(\mathcal{T}, q_{n+1}) := \\
&\notag\left[\hat{\model}^{{\text{lower}}}(q_{n+1}), \hat{\model}^{{\text{upper}}}(q_{n+1})\right],
\end{align}
where $\hat{\model}^{{\text{lower}}}$ and $\hat{\model}^{{\text{upper}}}$ denote the lower and upper quantile predictors, respectively. QR enables instance-dependent prediction intervals that adapt to heteroscedasticity and local uncertainty patterns.



\subsection{Implementation Details}
All experiments are conducted on a single NVIDIA A100 Tensor Core GPU.
Each experiment is repeated with 10 different global random seeds to ensure full reproducibility, including model initialization and data shuffling. 

Hyperparameters for each model–dataset configuration are independently tuned via grid search. We search over learning rates $\{0.0001, 0.001, 0.01\}$, embedding dimensions $\{64, 128, 256, 512\}$, and batch sizes $\{128, 256, 512, 1024, 2048, 4096\}$. 

Both variants of UKGE and PASSLEAF use early stopping based on the mean of validation loss and negative-sample validation loss, with patience set to 200 epochs due to slower convergence. BEUrRE, which converges more quickly, uses a shorter patience of 50 epochs. These values were empirically determined based on validation performance. Negative sampling ratios follow prior work: UKGE and PASSLEAF use 10 negative samples per positive triple, while BEUrRE, which benefits from higher negative pressure due to its probabilistic box structure, uses 30 negatives per positive. 

The final hyperparameters, selected using validation loss, are as follows: for UKGE, learning rate = 0.001, embedding dimension = 128, and batch size = 128 for CN15k and NL27k, or 256 for PPI5k. PASSLEAF consistently uses learning rate = 0.001, embedding dimension = 512, batch size = 512, and the Adam optimizer, with additional semi-supervised settings: $T_{\text{NEW\_SEMI}}=20$, $T_{\text{SEMI\_TRAIN}}=30$, $M_{\text{SEMI}}=0.8 \times$ batch size, sample pool size $C=10^7$, and $\alpha=0.02$. BEUrRE uses a learning rate of 0.0001, embedding dimension = 64, $\beta = 0.01$, and batch size = 4096 for CN15k, or 2048 for NL27k and PPI5k.

\section{Complete Results}\label{app: complete_results}
\begin{figure*}[h!]
    \centering
    \includegraphics[width=\textwidth]{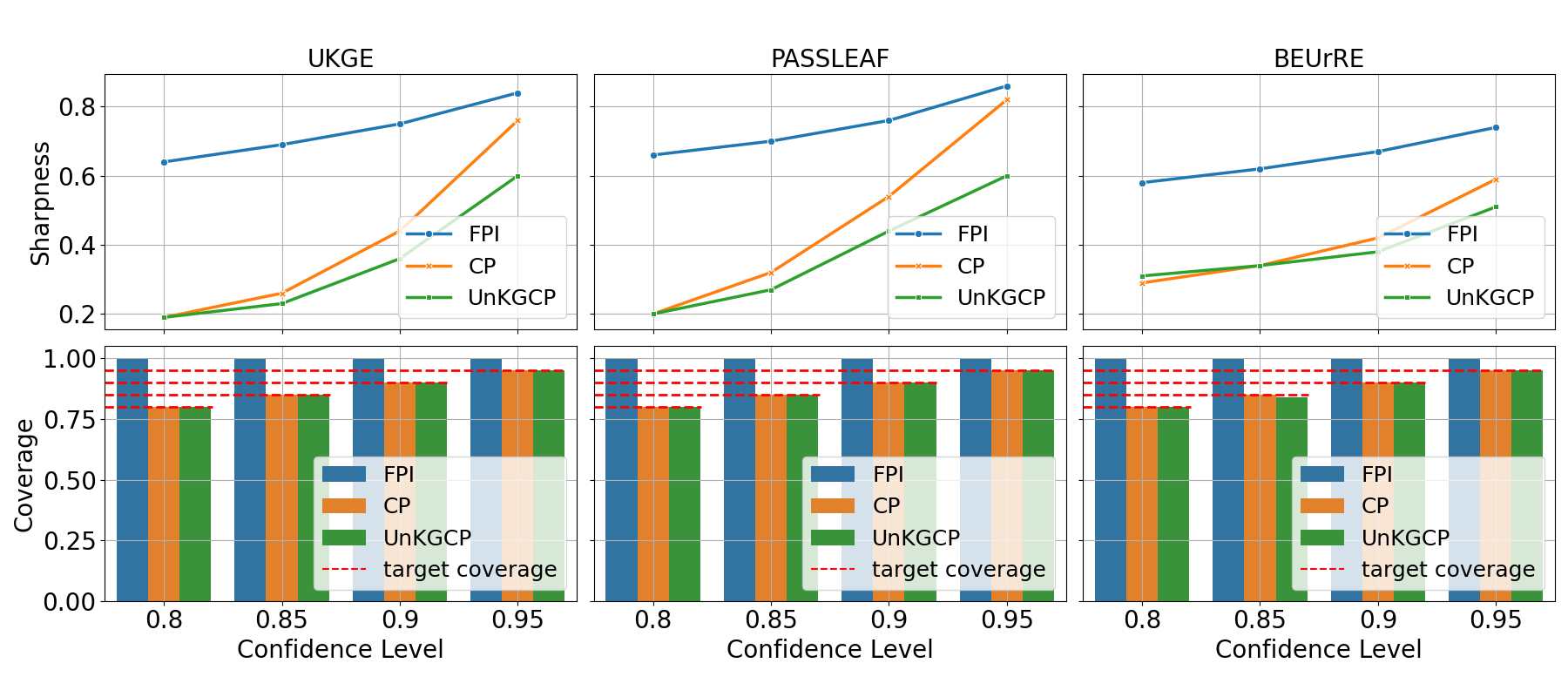}
    \caption{Effect of the confidence level $\alpha$ on the sharpness (top) and coverage (bottom) for \textbf{positive} test triples on NL27k. Each curve represents one predictor. Red dashed lines indicate the desired coverage levels.}
    \label{fig:confidence_level_NL27k_pos}
\end{figure*}

\begin{figure*}[h!]
    \centering
    \includegraphics[width=\textwidth]{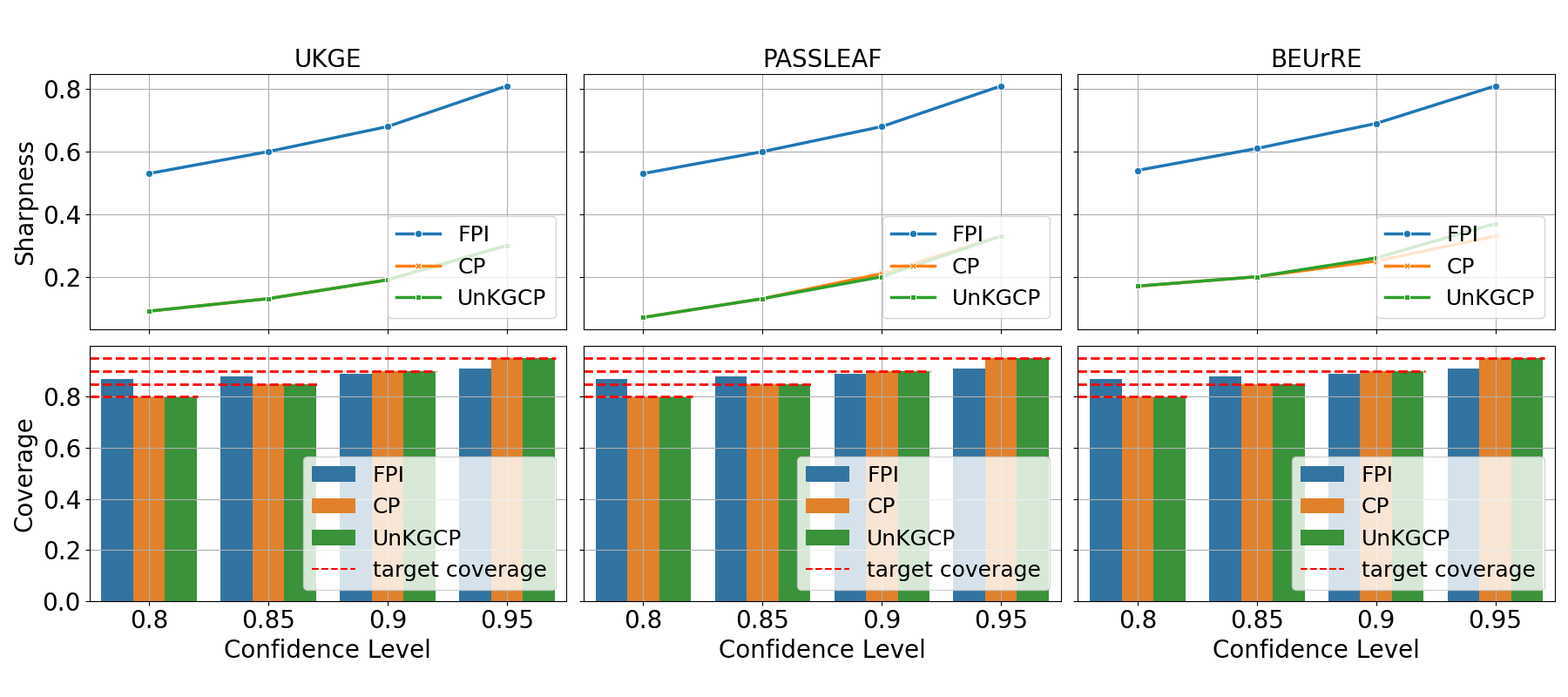}
    \caption{Effect of the confidence level $\alpha$ on the sharpness (top) and coverage (bottom) for \textbf{positive} test triples on PPI5k. Each curve represents one predictor. Red dashed lines indicate the desired coverage levels.}
    \label{fig:confidence_level_PPI5k_pos}
\end{figure*}

\begin{figure*}[h!]
    \centering
    \includegraphics[width=\textwidth]{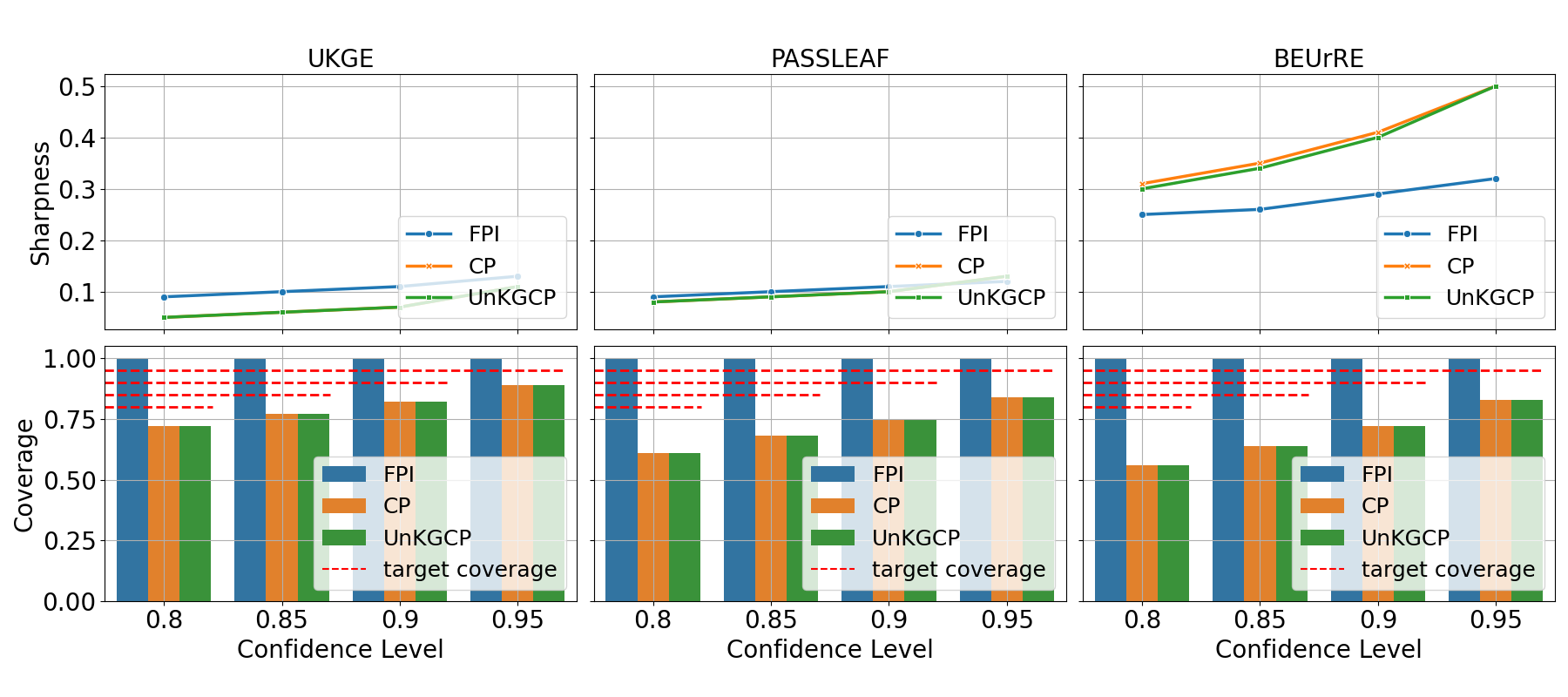}
    \caption{Effect of the confidence level $\alpha$ on the sharpness (top) and coverage (bottom) for \textbf{negative} test triples on CN15k. Each curve represents one predictor. Red dashed lines indicate the desired coverage levels.}
    \label{fig:confidence_level_CN15k_neg}
\end{figure*}

\begin{figure*}[h!]
    \centering
    \includegraphics[width=\textwidth]{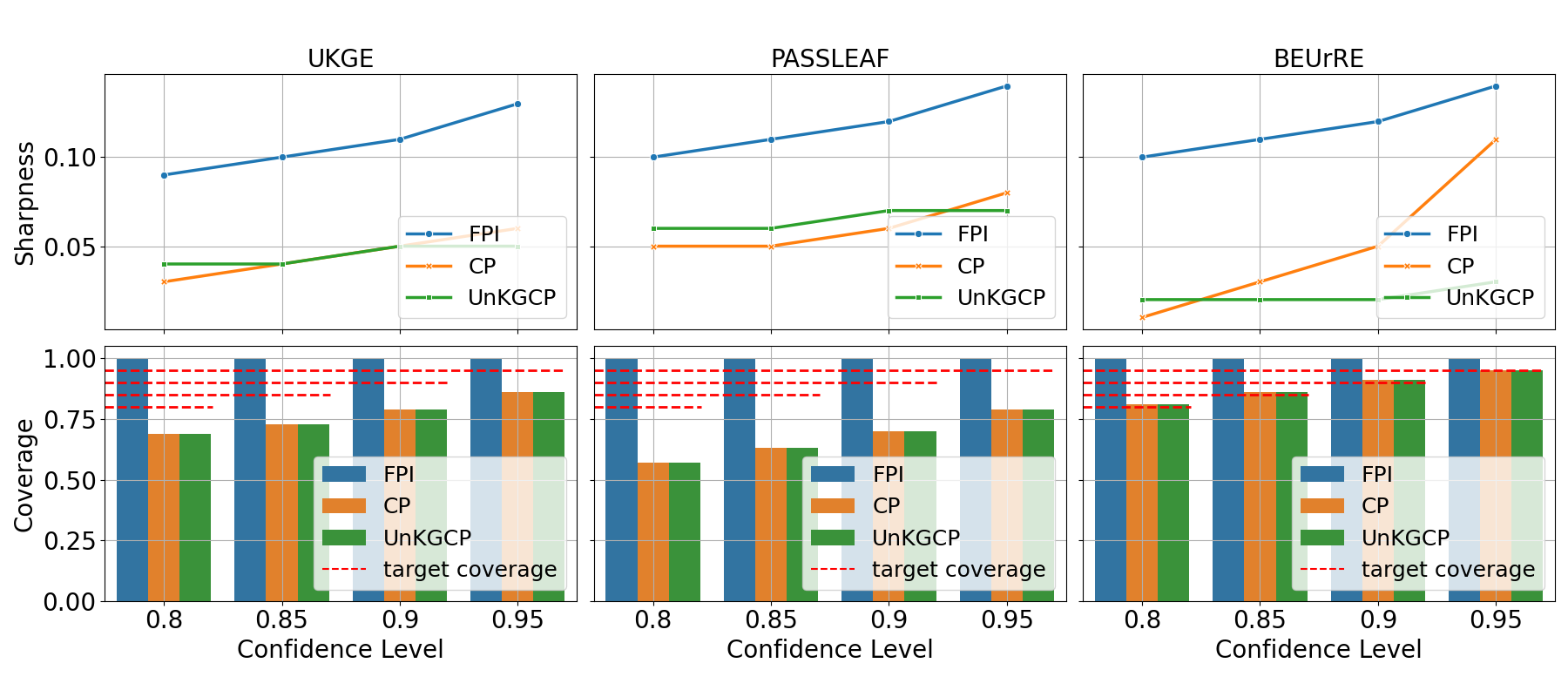}
    \caption{Effect of the confidence level $\alpha$ on the sharpness (top) and coverage (bottom) for \textbf{negative} test triples on NL27k. Each curve represents one predictor. Red dashed lines indicate the desired coverage levels.}
    \label{fig:confidence_level_NL27k_neg}
\end{figure*}

\begin{figure*}[h!]
    \centering
    \includegraphics[width=\textwidth]{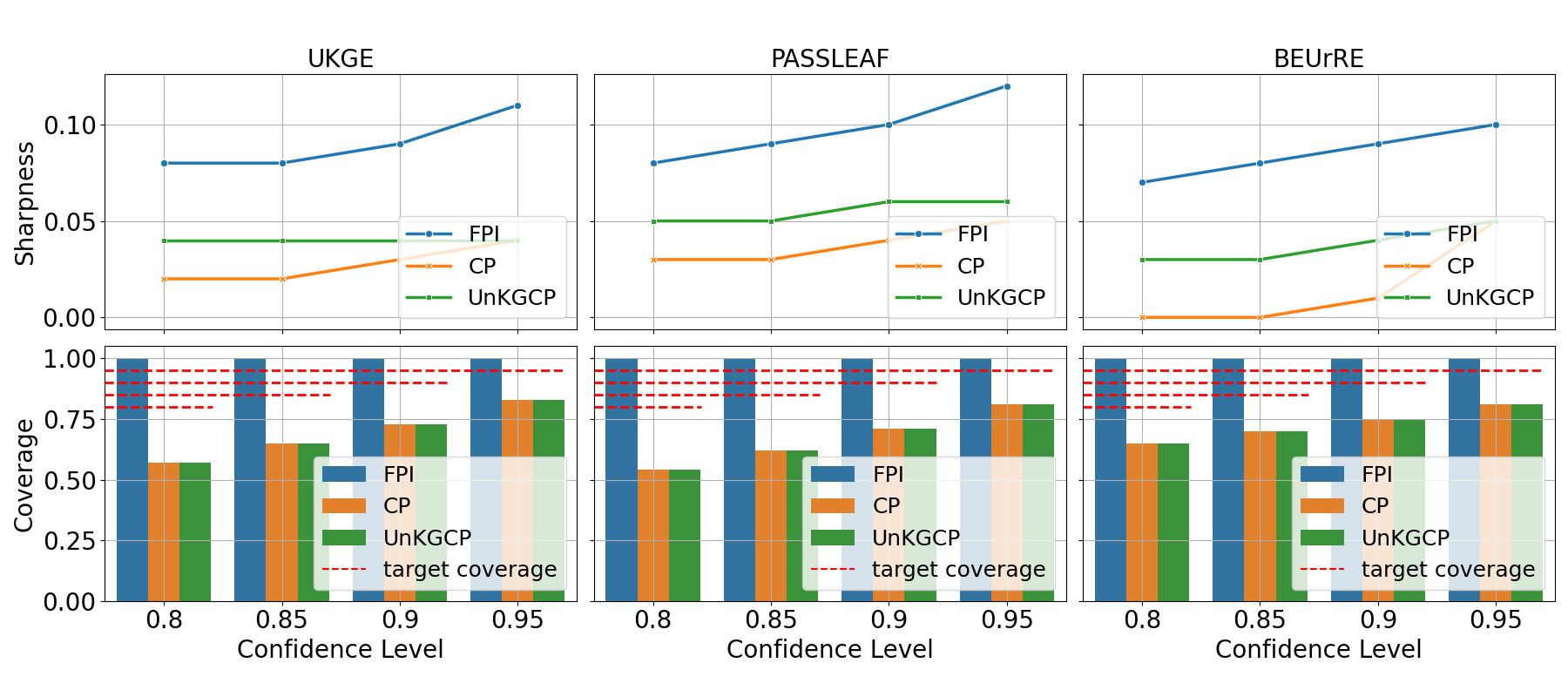}
    \caption{Effect of the confidence level $\alpha$ on the sharpness (top) and coverage (bottom) for \textbf{negative} test triples on PPI5k. Each curve represents one predictor. Red dashed lines indicate the desired coverage levels.}
    \label{fig:confidence_level_PPI5k_neg}
\end{figure*}

\begin{figure*}[t]
    \centering
    \includegraphics[width=\textwidth]{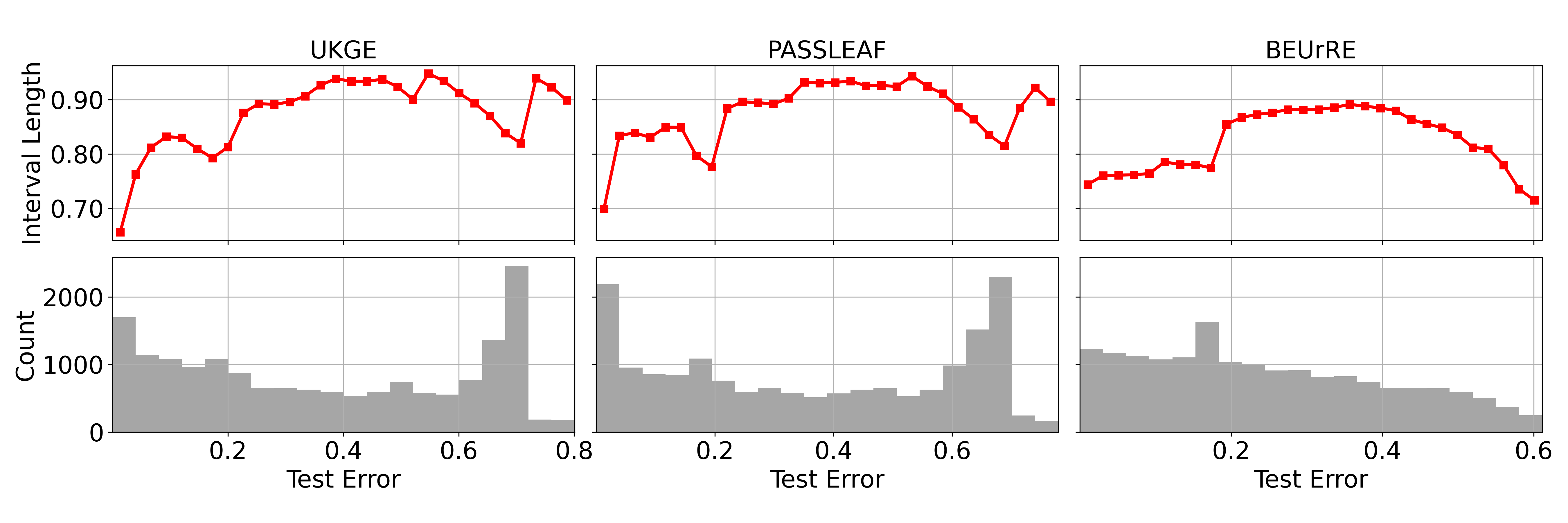}
    \caption{Conditionality analysis on CN15k (positive examples). }
    \label{fig:conditionality_cn15k}
\end{figure*}

\begin{figure*}[t]
    \centering
    \includegraphics[width=\textwidth]{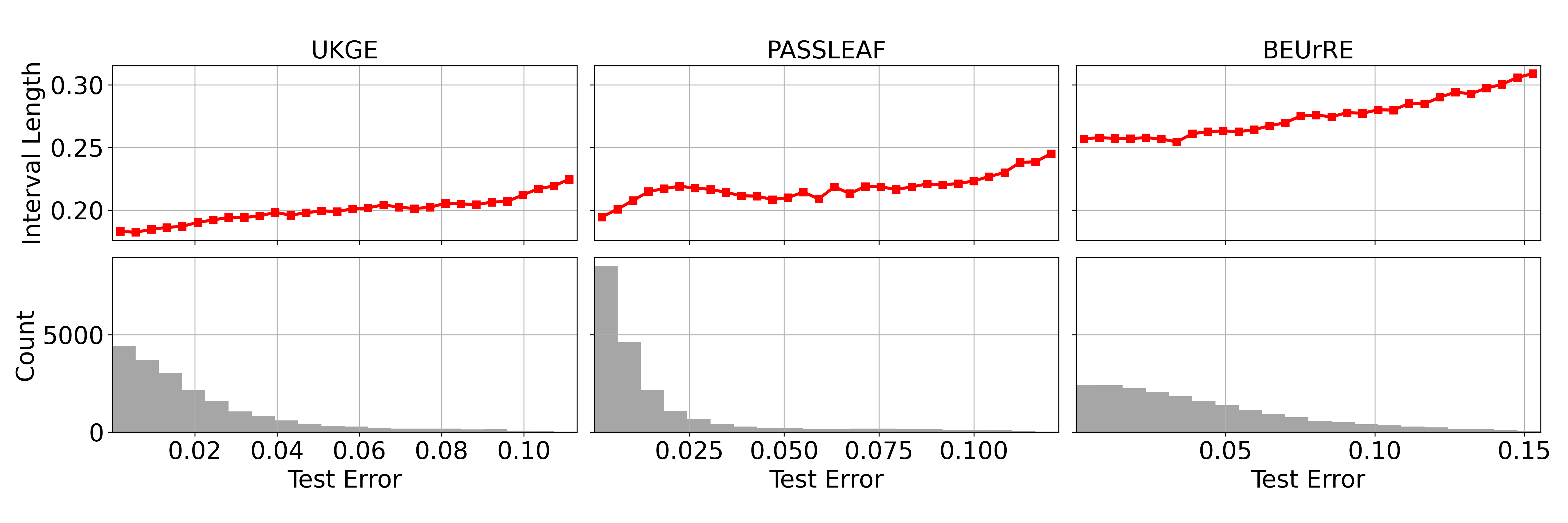}
    \caption{Conditionality analysis on PPI5k (positive examples). }
    \label{fig:conditionality_ppi5k}
\end{figure*}

\begin{figure*}[t]
    \centering
    \includegraphics[width=\textwidth]{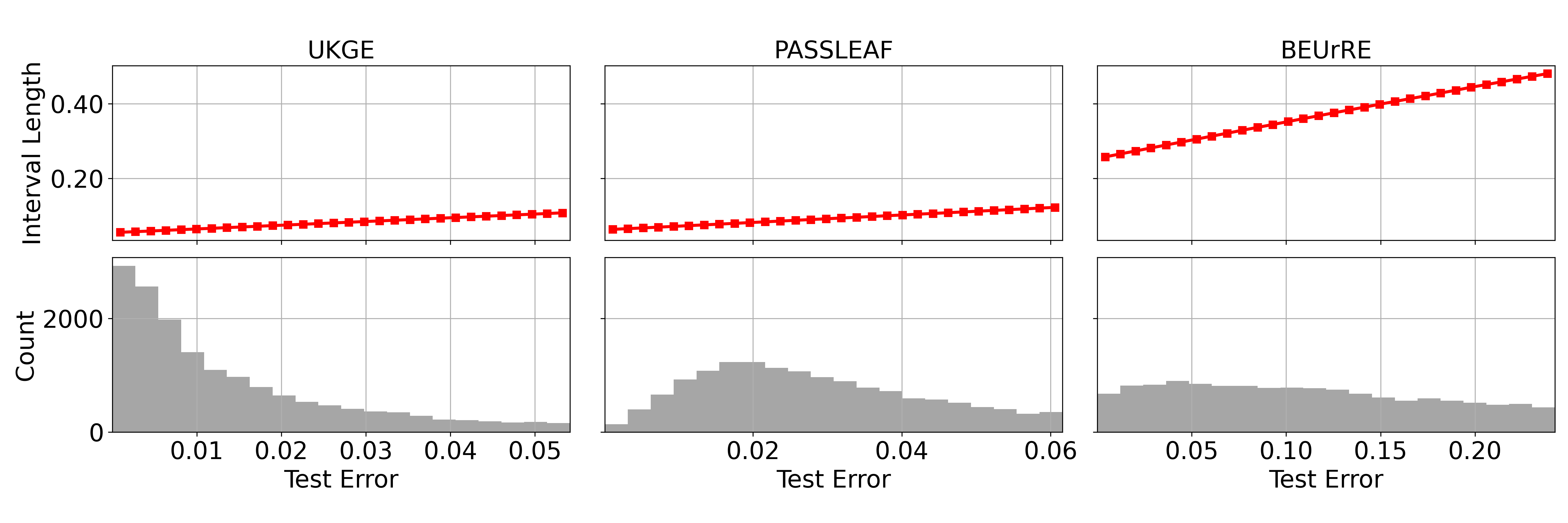}
    \caption{Conditionality analysis on CN15k (negative examples). }
    \label{fig:conditionality_cn15k_neg}
\end{figure*}

\begin{figure*}[t]
    \centering
    \includegraphics[width=\textwidth]{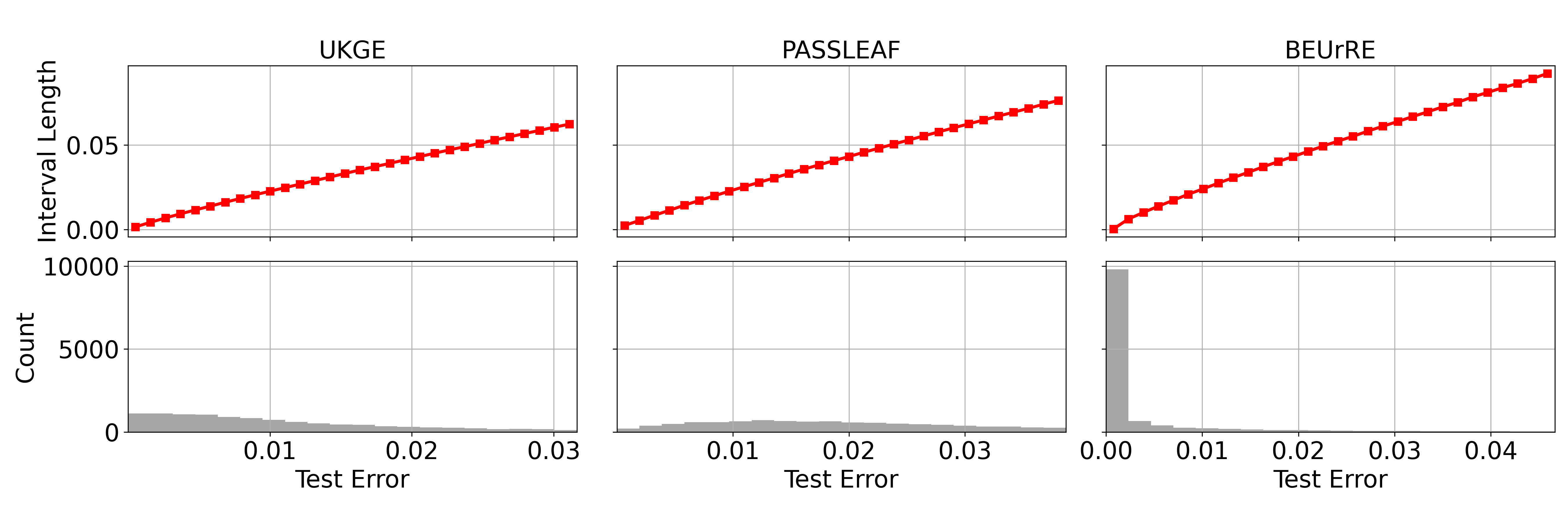}
    \caption{Conditionality analysis on NL27k (negative examples). }
    \label{fig:conditionality_nl27k_neg}
\end{figure*}

\begin{figure*}[t]
    \centering
    \includegraphics[width=\textwidth]{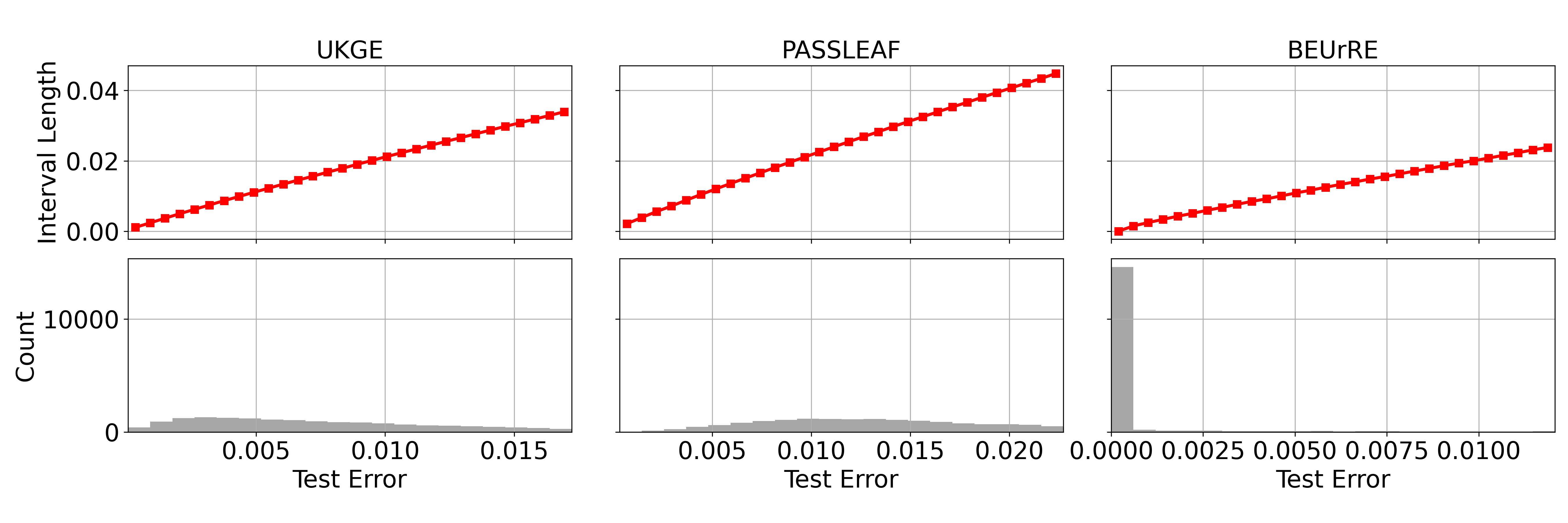}
    \caption{Conditionality analysis on PPI5k (negative examples). }
    \label{fig:conditionality_ppi5k_neg}
\end{figure*}

\begin{figure}[t]
    \centering
    \includegraphics[width=\linewidth]{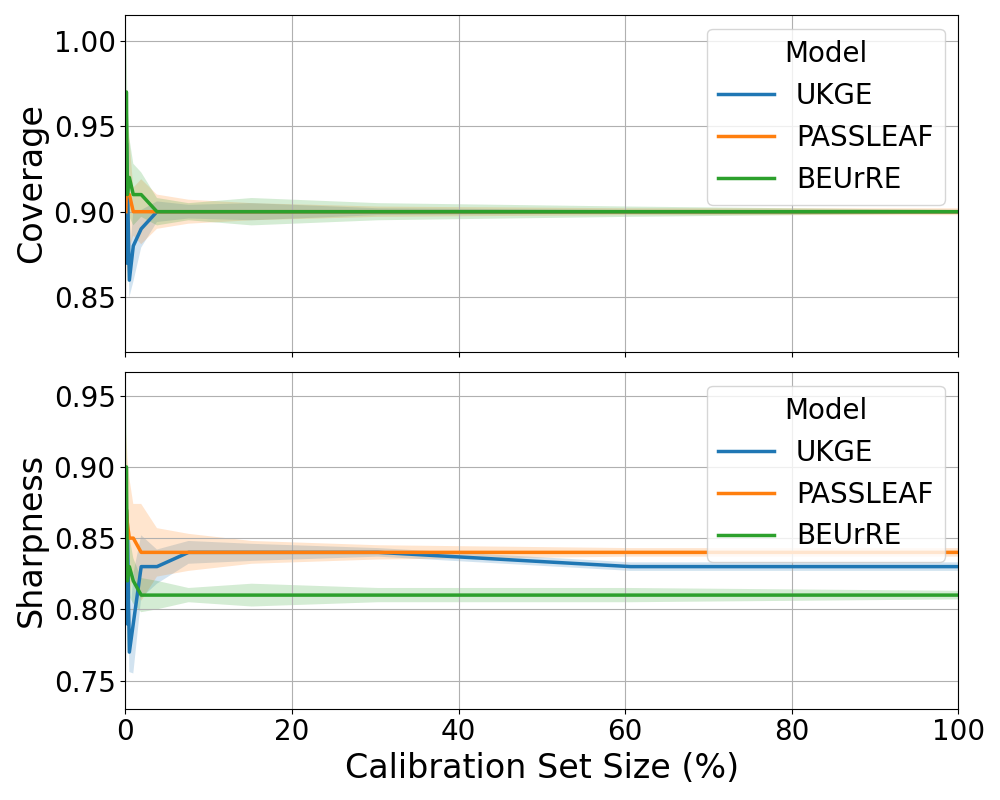}
    \caption{Effect of calibration set size on coverage and sharpness on CN15k (positive examples). The top panel reports coverage and the bottom panel reports sharpness. In both plots, the lines represent mean values across 10 runs, and the shaded areas indicate the standard deviation. }
    \label{fig:calib_size_analysis_cn15k_pos}
\end{figure}

\begin{figure}[t]
    \centering
    \includegraphics[width=\linewidth]{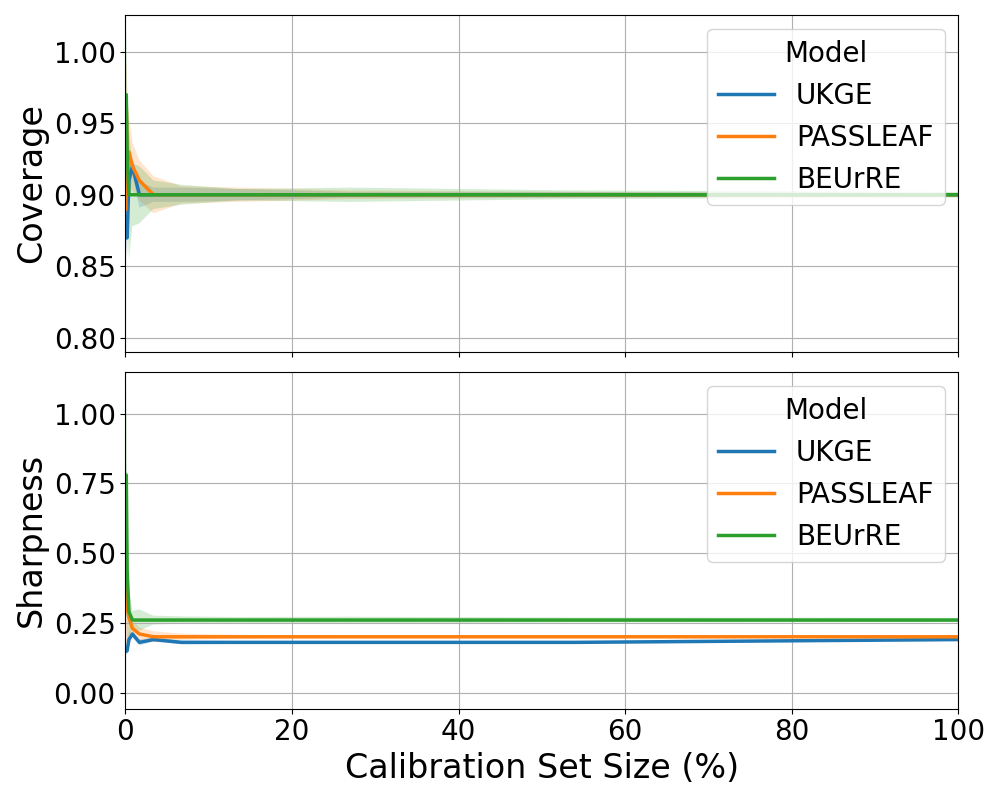}
    \caption{Effect of calibration set size on coverage and sharpness on PPI5k (positive examples). The top panel reports coverage and the bottom panel reports sharpness. In both plots, the lines represent mean values across 10 runs, and the shaded areas indicate the standard deviation. }
    \label{fig:calib_size_analysis_ppi5k_pos}
\end{figure}

\begin{figure}[t]
    \centering
    \includegraphics[width=\linewidth]{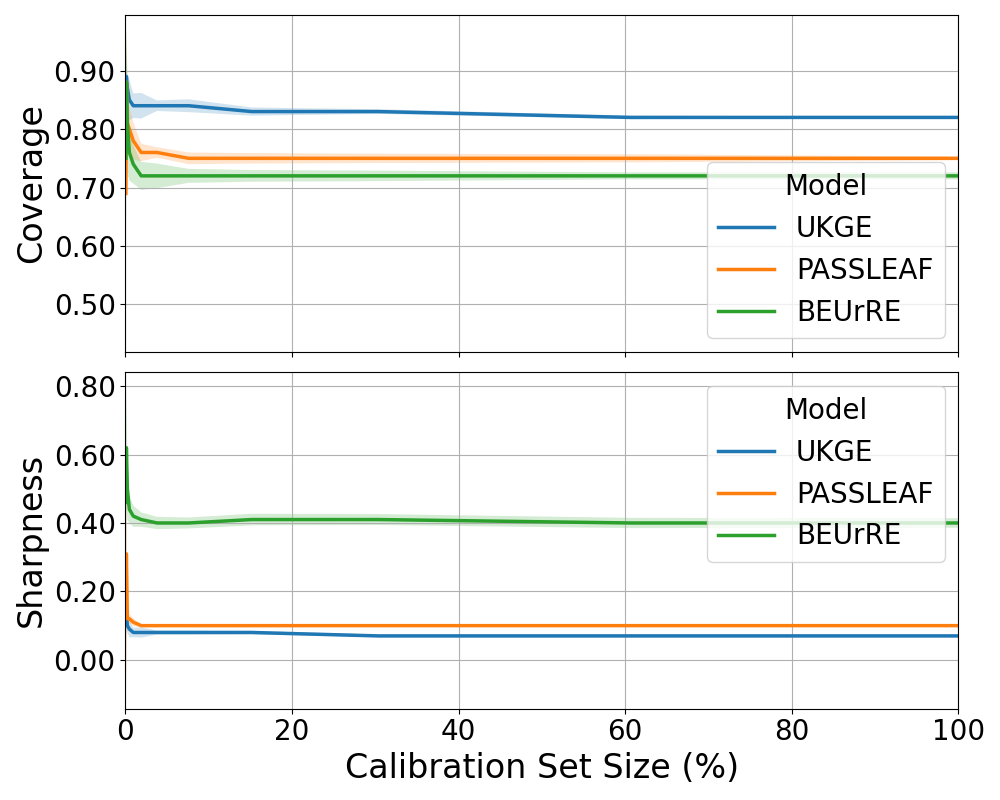}
    \caption{Effect of calibration set size on coverage and sharpness on CN15k (negative examples). The top panel reports coverage and the bottom panel reports sharpness. In both plots, the lines represent mean values across 10 runs, and the shaded areas indicate the standard deviation. }
    \label{fig:calib_size_analysis_cn15k_neg}
\end{figure}

\begin{figure}[t]
    \centering
    \includegraphics[width=\linewidth]{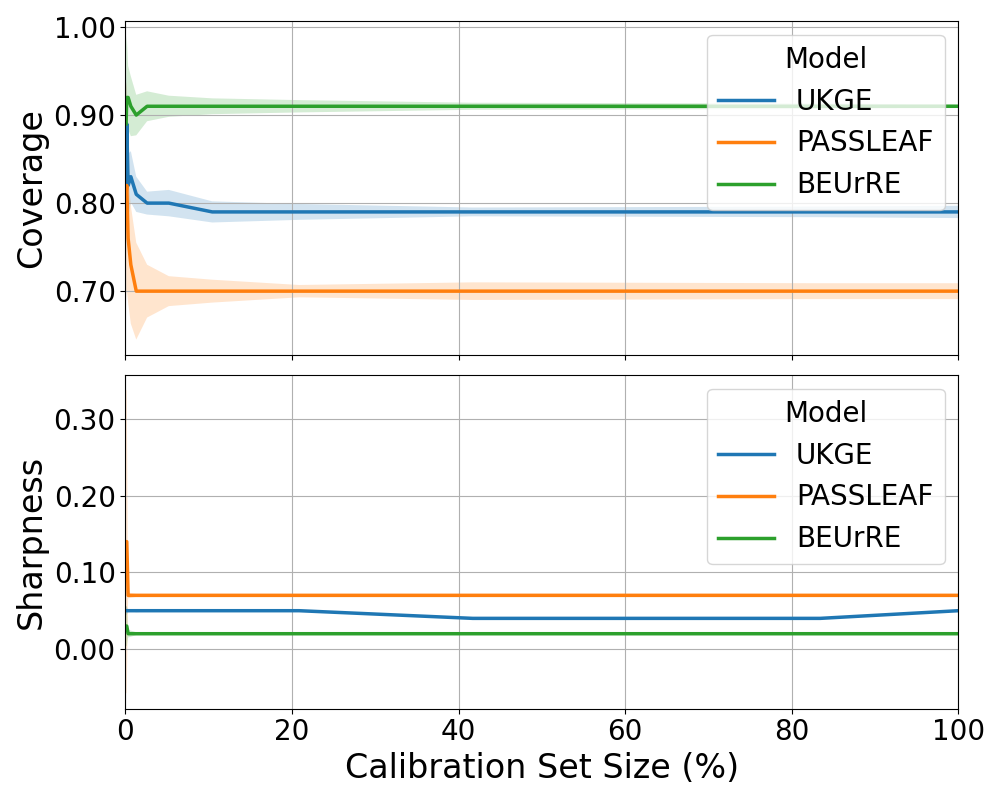}
    \caption{Effect of calibration set size on coverage and sharpness on NL27k (negative examples). The top panel reports coverage and the bottom panel reports sharpness. In both plots, the lines represent mean values across 10 runs, and the shaded areas indicate the standard deviation. }
    \label{fig:calib_size_analysis_nl27k_neg}
\end{figure}

\begin{figure}[t]
    \centering
    \includegraphics[width=\linewidth]{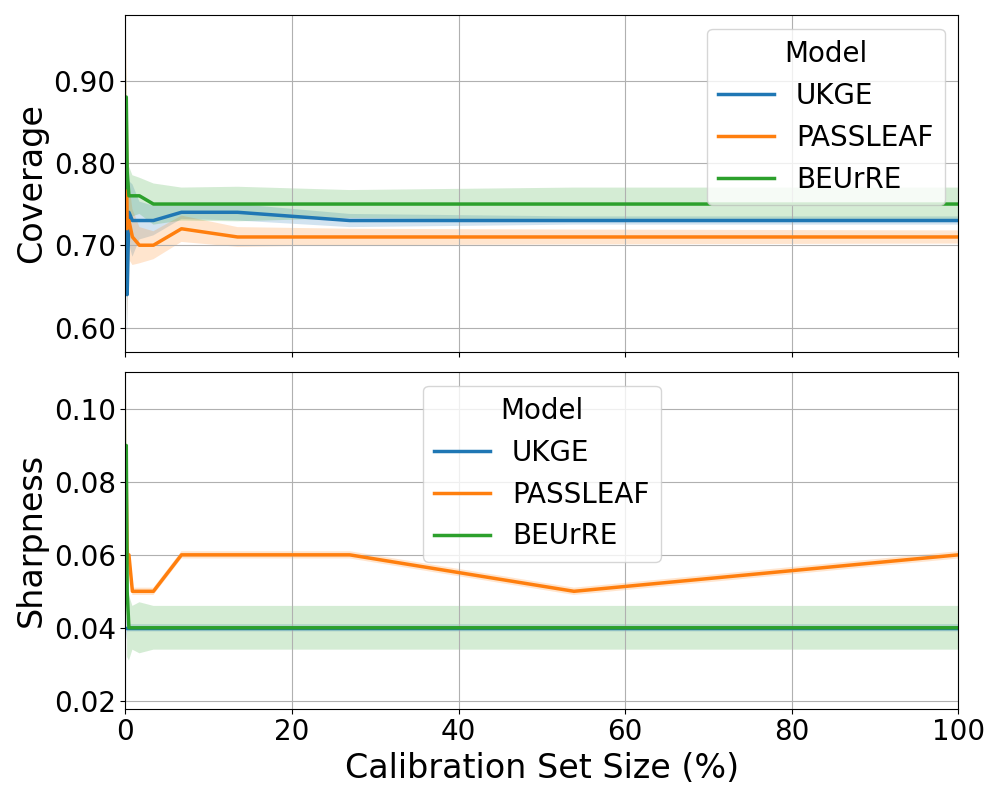}
    \caption{Effect of calibration set size on coverage and sharpness on PPI5k (negative examples). The top panel reports coverage and the bottom panel reports sharpness. In both plots, the lines represent mean values across 10 runs, and the shaded areas indicate the standard deviation. }
    \label{fig:calib_size_analysis_ppi5k_neg}
\end{figure}

For completeness, we provide the full set of experimental results, including those omitted from the main paper due to space constraints. 
This includes detailed coverage and sharpness values across all datasets, UnKGE models, and baselines (Figures~\ref{fig:confidence_level_NL27k_pos}–\ref{fig:confidence_level_PPI5k_pos}), 
as well as corresponding results on negative test triples (Figures~\ref{fig:confidence_level_CN15k_neg}–\ref{fig:confidence_level_PPI5k_neg}).
We also include comprehensive results for the conditionality analysis (Figures~\ref{fig:conditionality_cn15k}–\ref{fig:conditionality_ppi5k_neg}) and the sample efficiency analysis of the calibration step (Figures~\ref{fig:calib_size_analysis_cn15k_pos}–\ref{fig:calib_size_analysis_ppi5k_neg}). 
These results further support and strengthen the empirical findings presented in Section~\ref{sec:exp}.


\section{AI Assistants In Writing}
We use ChatGPT \cite{openai2024chatgpt} to enhance our writing skills, abstaining from its use in research and coding endeavors.





\end{document}